\documentclass[preprint,12pt]{elsarticle}

\usepackage{graphicx}
\usepackage{amssymb}
\usepackage{lineno}
\usepackage{amsmath}
\usepackage{tabulary}
\usepackage{ulem}
\usepackage{tabularx,ragged2e,booktabs,caption}
\usepackage{breqn}
\usepackage{siunitx}
\usepackage[inline]{enumitem}
\usepackage{xcolor}
\usepackage{mathrsfs}
\usepackage{hyperref}
\usepackage{bbm}
\usepackage{lipsum}
\usepackage{subfig}
\makeatletter
\def\ps@pprintTitle{%
 \let\@oddhead\@empty
 \let\@evenhead\@empty
 \def\@oddfoot{}%
 \let\@evenfoot\@oddfoot}
\makeatother

\begin{document}
\newcommand*{\estimates}{\mathrel{\hat=}}
\newcommand\independent{\protect\mathpalette{\protect\independenT}{\perp}}
\def\independenT#1#2{\mathrel{\rlap{$#1#2$}\mkern2mu{#1#2}}}

\begin{frontmatter}


\title{Spatiotemporal Adaptive Neural Network for Long-term Forecasting of Financial Time Series}


\author[label1]{Philippe Chatigny}
\ead{Philippe.Chatigny@usherbrooke.ca}
\author[label2]{Jean-Marc Patenaude}
\ead{jeanmarc@laplaceinsights.com}
\author[label1]{Shengrui Wang}
\ead{Shengrui.Wang@usherbrooke.ca}
\address[label1]{Department of Computer Science, Universit\'e de Sherbrooke, Sherbrooke, Qu\'ebec, Canada}
\address[label2]{Laplace Insights, Sherbrooke, Qu\'ebec Canada}
\begin{abstract}
Optimal decision-making in social settings is often based on forecasts from time series (TS) data. Recently, several approaches using deep neural networks (DNNs) such as recurrent neural networks (RNNs) have been introduced for TS forecasting and have shown promising results. However, the applicability of these approaches is being questioned for TS settings where there is a lack of quality training data and where the TS to forecast exhibit complex behaviors. Examples of such settings include financial TS forecasting, where producing accurate and consistent long-term forecasts is notoriously difficult. In this work, we investigate whether DNN-based models can be used to forecast these TS conjointly by learning a joint representation of the series instead of computing the forecast from the raw time-series representations. To this end, we make use of the dynamic factor graph (DFG) to build a multivariate autoregressive model. We investigate a common limitation of RNNs that rely on the DFG framework and propose a novel variable-length attention-based mechanism (ACTM) to address it. With ACTM, it is possible to vary the autoregressive order of a TS model over time and model a larger set of probability distributions than with previous approaches.  Using this mechanism, we propose a self-supervised DNN architecture for multivariate TS forecasting that learns and takes advantage of the relationships between them. We test our model on two datasets covering 19 years of investment fund activities. Our experimental results show that the proposed approach significantly outperforms typical DNN-based and statistical models at forecasting the 21-day price trajectory. We point out how improving forecasting accuracy and knowing which forecaster to use can improve the excess return of autonomous trading strategies.

\end{abstract}


\end{frontmatter}

\section{Introduction}
\label{S:1}

In recent decades, DNNs have improved TS forecast accuracy in various social settings ~\cite{makridakis2019forecasting}. Besides their ability to handle non-linear processes, they provide a cost-effective approach for uncovering relations between TS. DNNs are based on the dynamic factor graph (DFG) framework \cite{mirowski2009dynamic,mirowski2011time}, which is a particular case of a factor graph~\cite{koller2009probabilistic} in which the template method \cite{koller2009probabilistic} is applied. Specifically, a DNN-based model assumes that the factors of a DFG are individual neural networks (NN) that enforce a hierarchical structure for pattern detectors throughout its hidden layers \cite{bengio2009learning}. Under this framework, the DNN-based model learns complex probability distributions and increases forecast accuracy on mostly homogeneous datasets containing multiple measurements, as well as in applications where there are exogenous variables that are strongly related to the variable(s) of interest \cite{hyndman2018forecasting}: e.g., traffic \cite{yang2016optimized} or electricity load forecasting \cite{olagoke2016short}.

However, training a DNN remains difficult for most TS settings \cite{makridakis2018statistical, makridakis2019forecasting}, especially when TS are non-ergodic, heteroskedastic, non-stationary or have high noise-to-signal ratios. Such cases are often found in financial TS. Few DNN-based models have demonstrated consistent accuracy on such datasets spanning multiple years for different asset classes \cite{sezer2019financial}. Besides reasons associated with concept drift~\cite{vzliobaite2010learning}, the difficulties in forecasting financial TS are also due to the fact that most DNN learning frameworks do not appear to be adapted for this setting. Training a DNN needs a large dataset of independent training samples that are representative of the data to infer. Aside from applications like intra-day forecasting \cite{qin2017dual}, most financial applications rely on TS that have a relatively limited number of measurements \cite{makridakis2018statistical}. Additionally, historical price trajectories can be very noisy and their behaviors exhibit complex cyclical effects \cite{bruche2010recovery}. As it is not possible to obtain multiple independent realizations of a specific asset's price fluctuation under different circumstances for the same time period \cite{marshall2009principles}, the nature of financial TS necessarily leads to both a lack of training data and the well-known difficulty in modeling their long-term effects \cite{bengio1994learning}.

This paper proposes a more effective DNN framework for forecasting multiple financial assets conjointly and enhancing the capability of the TS model to learn a larger set of probability distributions. The key contributions of this paper are as follows:
\begin{enumerate}
 \item We propose a novel attention mechanism for the Dynamic Factor Graph (DFG) framework. This mechanism offers the capacity to consider a variable number of past latent states over time.
 
 \item We make use of the novel attention mechanism to optimize the order of an autoregressive (AR) generative function over time. We show how such a mechanism can model non-stationary distributions while keeping a constant parameterization.
 
 \item By incorporating the attention mechanism, we develop an energy-based deep generative approach for modeling interactions between multiple TS to generate multivariate forecasts. Our spatiotemporal adaptive neural network (STANN) is able to operate under a limited data constraint by exploiting prior knowledge of the TS to "virtually" augment its training samples and allows the discovery of interrelations between TS.

 \item We have conducted an extensive experimental evaluation showing the effectiveness of the proposed model for forecasting 21 daily return trajectories of exchange-traded funds (ETFs) and mutual funds (MFs). We also show preliminary but promising results of the proposed model for improving autonomous trading strategies. Of all the models proposed in the last 10 years \cite{sezer2019financial}, ours is, to our knowledge, the first to outperform naive baselines in a monthly multivariate financial TS setting.
 
\end{enumerate}
The remainder of this paper is organized as follows: Section~\ref{Related work} reviews major existing work on modeling TS in social settings and relevant notions related to the DFG. In Section~\ref{The STANN model}, we present our model and describe its training procedure. In Section~\ref{Experiments}, we present the setup of our empirical evaluation, which extends over more than 19 years of financial market activities, and describe our results. Section~\ref{Conclusion} presents our conclusion.

\section{Related Work}
\label{Related work}

\subsection{Prior Work:}

Different formulations \cite{rubanova2019latent,cho2014learning} of DNN models have been introduced to facilitate their application on TS data. While promising results have been achieved recently for financial TS prediction, as in \cite{borovykh2018dilated}, it has been pointed out that much of the published machine learning (ML) work in the TS literature claims satisfactory accuracy without adequately comparing the methods used against conventional methods \cite{makridakis2018statistical} and, further, that it relies on inappropriate criteria \cite{chatfield1988apples,hyndman2006another}. In fact, only a few authors, such as \cite{rangapuram2018deep,smyl_hybrid_2020}, have been able to show that their models yield better performance on multiple TS than simple statistical models like ARIMA or even a naive forecast. Most work uses non-scaled error metrics to assess forecast quality on multiple TS. However, it has been known for years that comparing forecasts of multiple TS of different scales via non-scaled metrics often leads to misleading results \cite{chatfield1988apples,hyndman2006another}. The myriad of existing DNN-based models \cite{sezer2019financial} applied to financial settings and the results presented around them have raised undue expectations that such methodologies provide accurate predictions at forecasting multiple TS, while there is clearly a lack of experimental demonstration that they outperform simple baselines in the majority of cases.

Nonetheless, large gains can still be achieved by using DNN and ML approaches. Recently, state-of-the-art accuracy was achieved at the M4 competition \cite{makridakis2018m4}, where the top 2 entries used DNN-based or ML techniques along with statistical models. Subsequent to these findings, the authors in \cite{oreshkin2019n} were the first to show that it was possible to build a pure DNN-based model for this task and achieve greater gains than the best competition entry \cite{smyl_hybrid_2020}. Given the wide range of TS to forecast\footnote{The M4 dataset contained 100,000 individual TS, of which approximately 25\% were financial TS of different types.}, the top-performing models submitted relied on ensemble techniques to be robust over the different types of series.

Direct comparison between single and ensemble models is generally unfair, as ensemble models permit the modeling of various probability distributions using multiple TS models and subsequently apply some form of forecast combination by evaluating the inference capabilities of each TS model a posteriori. However, the findings from these models can be investigated with a view to building better individual models. For instance, the DNN-based models, which performed well on this dataset \cite{smyl_hybrid_2020,oreshkin2019n}, provide insights into techniques that can be used to improve the performance of individual models: e.g., residual connections between hidden layers, adaptive learning rate scheduling, input preprocessing and both seasonal and trend decomposition embedded directly in the model. Most of these techniques are \textit{"tricks"} to facilitate DNN learning. However, the idea of applying a signal decomposition within a neural network is promising and several authors \cite{oreshkin2019n,godfrey2017neural,hansen2003forecasting} have shown its effectiveness on real-world datasets. Given the well-known difficulty of dealing with the raw signal of financial TS, we raise the question whether a better representation of these TS can be learned directly by applying such decomposition within the learned latent variables of a DFG \cite{mirowski2009dynamic}.

\subsection{Dynamic Factor Graph:}
\label{sub:Dynamic Factor Graph}

A DFG consists of an undirected acyclic state-space model where factors are replicated on a fixed time interval $T=\{t_1, ..., t_T\}$ to model a probability distribution. A DFG models the joint probability $P(X, Z;\textbf{W})$ of the observable values $X=\{x_1, ..., x_T\}$ and the latent variables $Z=\{z_1, ..., z_T\}$ given some parameterization of all factors in the graph $\textbf{W}$ as in Eq.~\ref{eq:dfg_joint}. Here, $\mathscr{L}$ is the partition function and $E(X,Z;\textbf{W}) \propto - logP(X,Z|W) + const$ is the \textit{total energy} of the model. The \textit{total energy} of the model is the sum of the normalized probability scalars assigned by a factor to all possible input data points associated with it. To obtain a probability, the total energy is normalized by $\mathscr{L}$.

\begin{equation}
\label{eq:dfg_joint}
    P(X, Z;\textbf{W}) = \frac{e^{-\beta E(X,Z;\textbf{W})}}{\int_{X'}\int_{Z'}e^{-\beta E(x',z';\textbf{W})}d x' d z'}  = \frac{e^{-\beta E(X,Z;\textbf{W})}}{\mathscr{L}}
\end{equation}
\begin{equation}
\label{eq:dfg_joint2}
         E(X,Z;\textbf{W}) = \sum_{t\in T}\sum_{F \in \mathscr{F}} E(A_t, O_t; F); A_t\in Z, O_t\in\{X, Z\} 
\end{equation}

\begin{equation}
\label{eq:dfg_joint3}
         E(A_t, O_t; F)  = \begin{cases}
         error(g(Z_t, \textbf{W}_{g}), Z_{t+1}) & \text{ if } F=g\\
         error(d(Z_t, \textbf{W}_{d}), X_t) & \text{ if } F=d
         \end{cases}
         ; \mathscr{F}=\{d,g\}
\end{equation}

The \textit{energy term} for a given sequence of observable values $X$ and latent states $Z$ is given by Eq.~\ref{eq:dfg_joint2}, with the energy term of a single factor defined by Eq.~\ref{eq:dfg_joint3}. Here we assume that our DFG follows a parameterization similar to that of an HMM architecture of order 1 with two factors, i.e., $\mathscr{F}=\{d,g\}$, and $\textbf{W}_F$ is the parameterization of the factor $F\in\mathscr{F}$. The higher the energy term between an input data point and its associated output data point, the less probable it is that the value will be observed. Despite the fact that the DFG's edges are undirected, the \textit{energy term} of each factor is not. Hence, training a DFG for TS forecasting is similar to adjusting the parameters of a Dynamic Bayesian Network (DBN) \cite{koller2009probabilistic} where we simply need to adjust the parameters of the factors using maximum likelihood estimation, which is equivalent to reducing the \textit{total energy} of the model.

For our particular case, where we consider an HMM under the DFG framework, the main difference between an RNN and this particular DFG is how the state-space component is used. However, since Eq.~\ref{eq:dfg_joint} is intractable for continuous variables under non-Gaussian distributions, we estimate the mode of the distribution instead by maximum a posteriori approximation \cite{mirowski2011time}. Thus, an HMM-based DFG model learns the probability distribution $\mathbb{P}$ of a TS using two factors replicated over time: a decoder factor and a dynamic factor, i.e. $\mathscr{F}=\{d, g\}$. The decoder factor \(d(Z;\textbf{W}_d)\) is a function that models the maximum likelihood of observing a random variable $X_t$ given latent variable $Z_t$: 

\begin{equation}
    \tilde{X_{t}} = d(Z_{t}, \textbf{W}_d) \estimates \arg\max_{x} \mathcal{L}(x|Z_t=z_i; \textbf{W}_d); z_i \in Z, x \in X 
    \label{eq:decoder_factor}
\end{equation}

\noindent with $\textbf{W}_d$ being the parameterization of the factor, $t$ a particular time point and $\mathcal{L}$ the likelihood function. The dynamic factor $g(Z;\textbf{W}_g)$ models the maximum likelihood of observing a state given some prior state and is defined by Eq.~\ref{eq:transition_factor}. $g(Z;\textbf{W}_g)$ models a transition probability distribution as in the DBN framework:

\begin{equation}
\begin{split}
    Z_{t+1} = g(Z_{t}, \textbf{W}_g) \estimates \arg\max_{z} \mathcal{L}(Z_{t+1}=z|Z_{t}=z_j, \cdots, Z_{t-k}=z_{j'});\\ z, z_j, z_{j'} \in Z
\end{split}    
\label{eq:transition_factor}
\end{equation}

Note that one must specify the order of \(g(Z;\textbf{W}_g)\) by changing its configuration: \(Z_{t+1} = g(Z_{t}, ..., Z_{t-k}; \textbf{W}_g))\), where $k$ is the autoregressive (AR) order of the process. Doing so makes the assumption that the probability distribution $\mathbb{P}$ models a stationary process if, for all $t$, $\mathcal{L}(Z_{t+1}=z_i|Z_{t-k:t}=z_j)\geq 0$ and is constant for all $t$ \cite{koller2009probabilistic}\footnote{Here, $Z_{t-k:t}$ corresponds to all $Z_{t'}$  included between $Z_{t-k}$ and $Z_{t}$, where $t'\in T$}. This assumption holds for both the discrete and the continuous case \cite{keener1993perron,glynn2018probabilistic}. When modeling $\mathbb{P}$ using a graphical model, we use the notation $\mathbb{P} \models (A\independent B | C)$~\cite{koller2009probabilistic} to indicate that $\mathbb{P}$ models a local independence relation between a set of nodes $A$ and $B$ given $C$. In particular, assuming that the probability distribution models a stationary process induces the following set of independence relations:
\begin{itemize}
    \item The latent variable evolves in a \textit{Markovian} or a \textit{semi-Markovian} way:
    \begin{equation}
        \mathbb{P} \models (Z_{t+1} \independent Z_{0:t-1-k}| Z_{t-k:t})
        \label{eq:stationnay_dep1}
    \end{equation}
    \item The observation variables at time t are conditionally independent of the state sequence given the latest $k+1$ state variables at time $t$ \cite{koller2009probabilistic}: 
    \begin{equation}
        \mathbb{P} \models (X_{t} \independent Z_{0:t-1-k}, Z_{0:t+1:\infty}| Z_{t-k:t})
        \label{eq:stationnay_dep2}
    \end{equation}
\end{itemize} 

This AR order is a hyperparameter that needs to be tuned carefully, since the true probability distribution is intractable in most cases \cite{mirowski2011time} and the set of local independences cannot be verified in practice. Assuming that the AR parameters are constant can impair training, as the resulting AR weights are optimized to reduce the average error. This limitation is problematic if the AR order was not selected appropriately or the training data contains multiple TS dynamics. In this work, we address these limitations by proposing an attention mechanism that enables a DFG to select its AR order automatically and adjust it over time. The stationary assumption can thus be relaxed such that the set of interdependences in Eq.\ref{eq:stationnay_dep1} and Eq.\ref{eq:stationnay_dep2} holds but the process order $k$ is a function of time. This permits non-stationary probability distributions to be modeled since $\mathcal{L}(Z_{t+1}=z_i|Z_{t-k:t}; \textbf{W}_g)\geq 0$ but is not necessarily constant over time.

\begin{figure}[ht]
\label{fig:dfg}
\centering 
\includegraphics[width =0.80\textwidth]{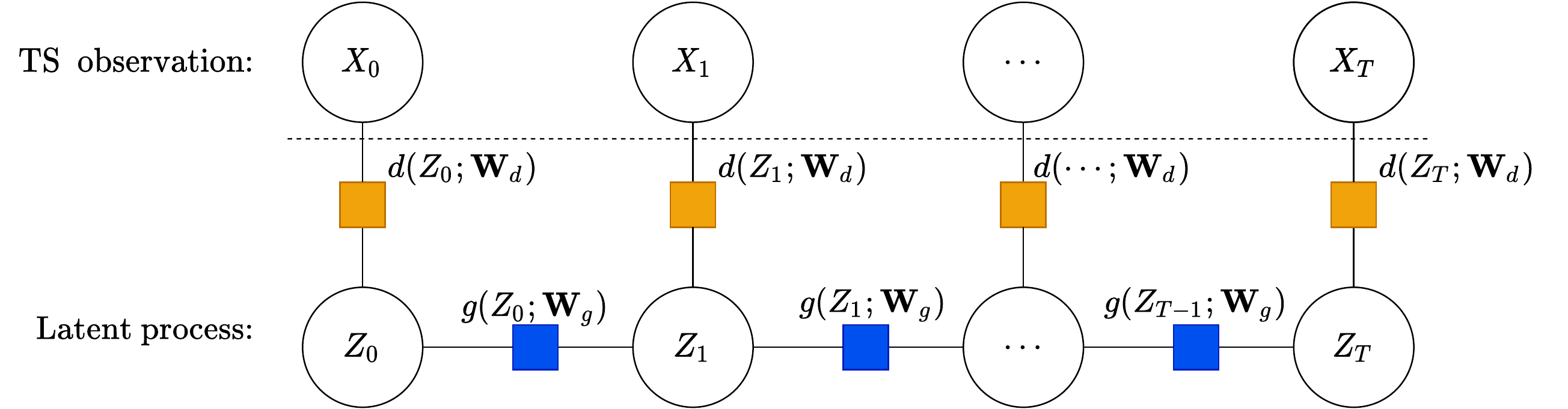}
\caption{An HMM-based DFG architecture that admits observed variables $X$ and latent variable $Z$. Both decoder (orange squares) and dynamic factors (blue squares) can be implemented as parametric functions and be trained using gradient descent. Notice that the dynamic process of the series is captured entirely in the latent space: \(Z_{t+1} = g(Z_{t}; \textbf{W}_g)\). Thus, an HMM-based DFG is a particular case of an RNN where the hidden states are directly learned instead of being computed explicitly by a function of past inputs.}
\end{figure}

\section{The STANN model}
\label{The STANN model}

\subsection{Model Definition Without Including Time Series Dependencies}
Given \(X: \mathbb{R}^{T \times n \times m}\), a 3-dimensional tensor representing a set of \(n\) TS of length \(T\) and dimensionality \(m\), we define \(X_{t, i, j}\) as the value of dimension \(j\) for TS \(i\) at time \(t\). The task of interest is to predict \(n\) multivariate TS \(\tau\) time steps ahead \(\tilde{X}: \mathbb{R}^{\tau \times n \times m}\). We represent the spatial relationship between series within a 3-dimensional tensor, that we denoted by \(W: \mathbb{R}^{n \times R \times n}\), where \(R\) is the number of relations considered. Thus, our aim is to train a model \(\textit{f}: \mathbb{R}^{T \times n \times m} + [\mathbb{R}^{n \times R \times n}] \rightarrow \mathbb{R}^{\tau \times n \times m}\). 

In STANN, we use a particular formulation of the DFG. The \textit{decoder} factor \(d(Z; \textbf{W}_d)\) decodes the expected variation between \(X_{t-1}\) and  \(X_{t}\) from the latent factor \(Z_t\), which allows the decoder to be defined as in Eq.~\ref{eq:decoder}. Here \(\tilde{X_t}\) is the prediction computed at time \(t\).

\begin{equation}
\label{eq:decoder}
\tilde{X}_t = X_{t-1} + d(Z_{t}; \textbf{W}_d) 
\end{equation} 

The dynamical module \(g(Z; \textbf{W}_g)\) is defined by Eq.~\ref{eq:dynamic} and considers the past \(k+1\) relevant latent factors \(Z_{t - k: t}\), i.e. \(Z_{t - k }\) to \(Z_{t}\). \(d(Z;\textbf{W}_d)\) and \(g(Z;\textbf{W}_g)\) is implemented as a doubly residual stacking NN, as in N-BEATS \cite{oreshkin2019n}. In contrast to N-BEATS, we apply the TS decomposition on the latent factors rather than the raw signals.

\begin{equation}
\label{eq:dynamic}
\tilde{Z}_{t+1} = g(Z_{t - k:t}; \textbf{W}_g)
\end{equation}

\subsection{Adaptive Computation Time for Autoregressive Order Selection}
\label{Model definition without spatial dependencies}

As mentioned in the previous section, assuming that the forecast depends on a fixed AR order covering the past \(k\) observations is a strong assumption that can impair model training if the autoregressive order is not selected correctly. RNNs, like the long short-term memory (LSTM) nework~\cite{hochreiter1997long}, consider the past $k$ observations by maintaining in memory a state vector that allows them to retain information as long as required and forget it when it is no longer relevant. Unlike LSTM, we permit \(k\) to vary adaptively without requiring the observable value as input. To this end, we propose an adaptive attention-based mechanism to enable DFG to be \textit{memory-augmented}. Our attention mechanism is inspired by the \textit{Adaptive Computation Time} (ACT) algorithm proposed in \cite{graves2016adaptive}, denoted as \(actm(Z; \textbf{W}_{actm})\) in our model. \(actm(Z; \textbf{W}_{actm})\) is a factor that generates a probability distribution on \(Z\) that is used to select the order of the regression in $Z_{t - k:t}$. The probability function associated with the factor at each time $t$ is modeled by a parametric function $ f_{actm}(Z_{t}; \textbf{W}_{actm})$, with $\textbf{W}_{actm}$ being its parameterization. For ease of computation, the order selection and the computation of \(Z_{t - k: t}\) are performed by using the sum of a certain number of past latent factors weighted by probability distribution function $f_{actm}$ as in Eq.~\ref{eq:actm}. 

\begin{subequations}
\begin{equation}
\label{eq:actm}
Z_{t-k:t}=actm(Z_t, \textbf{W}_{actm})=\sum_{0\leq k'<k:[b(t)>\epsilon]}\varphi_{Z_{t-k'}}Z_{t-k'}
\end{equation}
\begin{equation}
     \varphi_{Z_{t-k'}} = \begin{cases}  f_{actm}(Z_{t-k'}, \textbf{W}_{actm}) & \text{if } b(t)-f_{actm}(Z_{t-k'}, \textbf{W}_{actm}) > \epsilon \\
     b_{\texttt{cost}} & \text{if }  b(t)-f_{actm}(Z_{t-k'}, \textbf{W}_{actm}) \leq \epsilon \\
     \end{cases}
\end{equation}
\end{subequations}

Specifically, \(actm(Z;\textbf{W}_{actm})\) is implemented with two budgets \(b(t) = \{b_{\texttt{time}}=t,b_{\texttt{cost}}=1\}\): one to keep account of available past time steps and one to track the cost of considering a latent factor. We start with the current latent state factor \(Z_{t}\) and consider whether to include \(Z_{t-k'}\) for $k'=0, 1, 2, ...$.  Each time we consider a latent factor \(Z_{t-k'}\), we reduce our budget \(b_{\texttt{time}}\) by 1 and \(b_{\texttt{cost}}\) by \(\varphi_{Z_{t-k'}}=f_{actm}(Z_{t-k'}, \textbf{W}_{actm})\), the latter being bounded within \(]0, 1[\). If a budget goes below \(\epsilon\), i.e., either \(b_{\texttt{cost}}<\kappa\) or \(b_{\texttt{time}}=0\), we stop considering any more latent factors and attribute the remaining cost budget to the last factor considered. \(\kappa\in\mathbb{R}^+\) is a small constant (0.01 for the experiments in this paper), whose purpose is to allow the selection of an AR(1) process.

\begin{figure}[htbp]
\centering 
\includegraphics[width =0.95\textwidth]{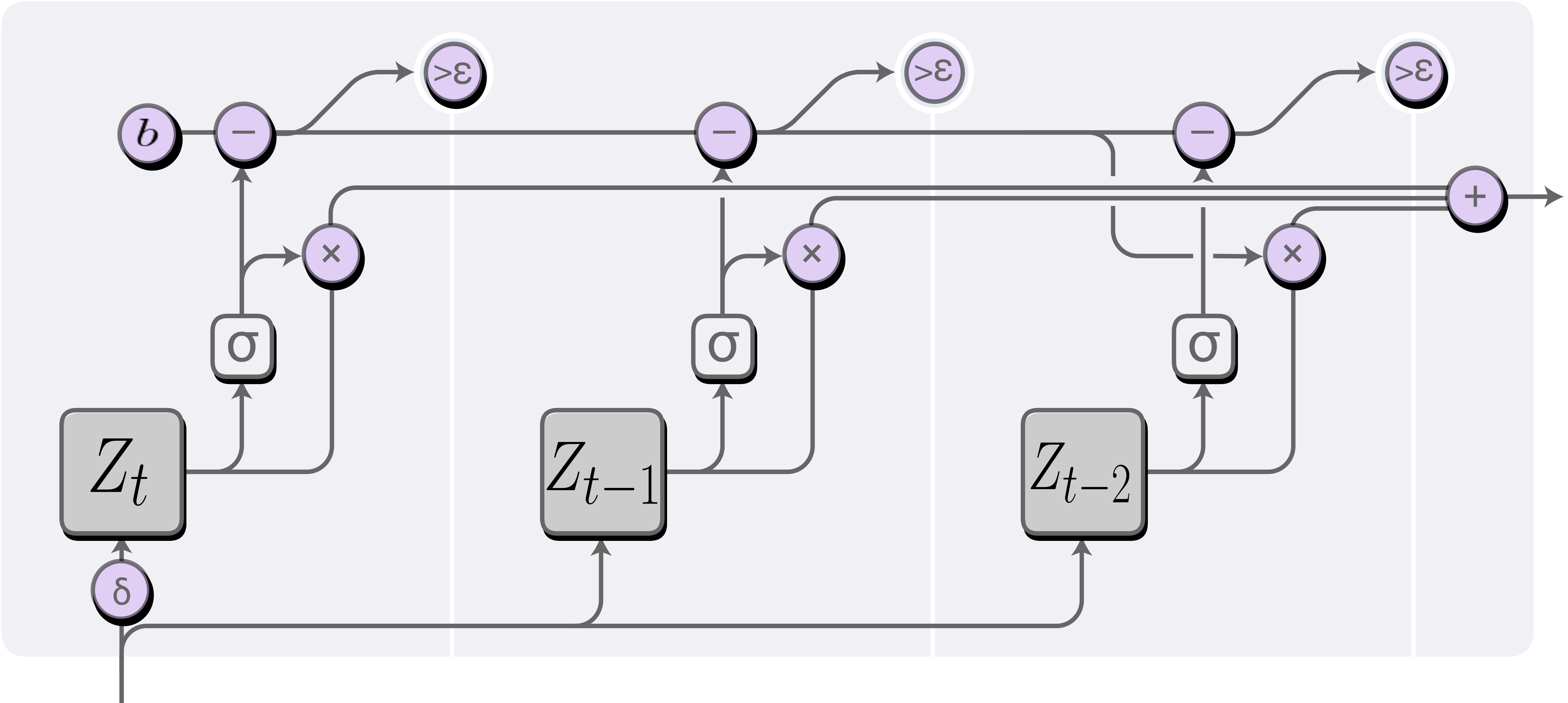}
\captionof{figure}{Illustration of the proposed attention mechanism. For illustration purposes, \(b\) includes both \(b_{\texttt{time}}\) and \(b_{\texttt{cost}}\), and \(\sigma\) denotes the AR weight produced by \(actm(Z_{t-i})\). The drawing was adapted from \protect{\cite{olah2016attention}}.
}
\label{fig:adaptive_rnn}
\end{figure}

\begin{figure}[htbp]
\centering 
\includegraphics[width =0.95\textwidth]{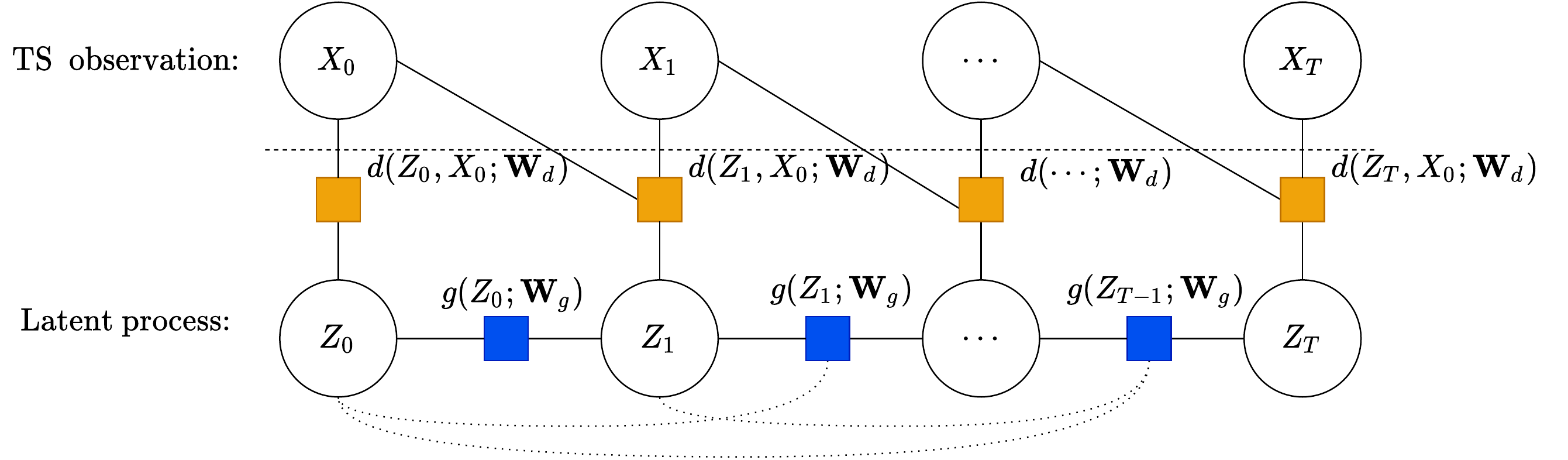}
\caption{Illustration of our model. The dotted lines at the bottom of the graph represent the possible relations between variables and the dynamic factors that $actm(Z;\textbf{W}_{actm})$ can consider. For illustration purposes, $actm(Z;\textbf{W}_{actm})$ and $g(Z;\textbf{W}_{g})$ are represented by the same factor (blue square in this figure).}
\label{fig:stann_dfg}
\end{figure}

Hence Eq.~\ref{eq:dynamic} can be reformulated as Eq.~\ref{eq:remormulation_dynamic}.
\begin{equation}
    \tilde{Z}_{t+1} = g(Z_{t - k:t}; \textbf{W}_g, \textbf{W}_{actm})
\label{eq:remormulation_dynamic}
\end{equation}
We can interpret \(actm(Z; \textbf{W}_{actm})\)'s objective as evaluating the quality of each past latent factor and assigning the appropriate autoregressive  weight at times \(t-k'\) that maximizes the log likelihood of the generative process modeled by Eq.~\ref{eq:dynamic}. Since \(actm(Z; \textbf{W}_{actm})\) uses \(b_{\texttt{time}}\) to determine how many past steps are available, we can theoretically account for all previous learned factors if \(\sum_{k=0}^{t} f_{actm}(Z_{t-k}, \textbf{W}_{actm}) < 1-\kappa\). Note that the imposed budget restricts each autoregressive weight to be between \(0\) and \(1\), with the sums of all the weights being equal to \(1\). We apply this mechanism solely within \(g(Z; \textbf{W}_g)\) to facilitate the training model, but the approach could also be extended to $d(Z;\textbf{W}_d)$. From now on, then, we will simplify our notation by considering that $\textbf{W}_g$ also includes the ACTM parameters. The attention mechanism is summarized in Figure~\ref{fig:adaptive_rnn} and can be designed as any configuration of a feedforward network with a sigmoid activation function. An illustration of our model with ACTM is presented in Fig.~\ref{fig:stann_dfg}.

The training procedure consists of minimizing the following bi-objectives loss function~(\ref{loss_dfg}):
    
    \begin{subequations}
    \begin{equation}
    \label{loss_dfg1}
        Loss(d, g, Z) = + {\dfrac{1}{T}} \sum_{t=1}^{T-1}\Delta(X_{t-1} + d(Z_{t}; \textbf{W}_d), X_t)
    \end{equation}
    \begin{equation}
    \label{loss_dfg2}
    + \lambda {\dfrac{1}{T}} \sum_{t=1}^{T-1} || Z_{t+1} - g(Z_{t - k:t}; \textbf{W}_g))||^{2}
    \end{equation}
    \label{loss_dfg}
    \end{subequations}
The first term~(\ref{loss_dfg1}) measures the ability of the model to reconstruct \(X_t\) from \(Z_t\). The second term~(\ref{loss_dfg2}) measures the system’s capacity to capture the dynamicity of the equation by its ability to link states of \(Z\) in sequential order. \(\Delta\) is a loss function that measures the difference between the prediction \(\tilde{X_t}\) and the ground truth \(X_t\). The second term in Eq.~\ref{loss_dfg2} forces the model to learn latent factors $Z_{t+1}$ that are as close as possible to $g(Z_{t - k:t}, \textbf{W}_g)$, with the ideal case being a solution where $Z_{t+1} = g(Z_{t - k:t}, \textbf{W}_g)$. However, this solution is not valid for the first term in Eq.~\ref{loss_dfg1}, where the ideal case is a solution for which $X_{t+1} = d(Z_{t - k:t}, \textbf{W}_d)$. To balance the relative importance of the two terms, the hyperparameter $\lambda$ is introduced to reduce or increase the importance of the second term relative to the first term in Eq.~\ref{loss_dfg1}. Training the model using Eq.~\ref{loss_dfg}\footnote{In our experiments, $\lambda$ was fixed trough a hyperparameter optimization that seach the optimal value in the following interval [0.01 and 1.0]} can be accomplished using any expectation-minimization-based approach \cite{mirowski2009dynamic} or an end-to-end \cite{tesauro1995temporal} approach that trains the three factors conjointly.

\subsection{Model Definition Including Time Series Dependencies}
\label{Model definition including spatial dependencies}

Let us now introduce the way interrelations between TS are captured. As pointed out in \cite{schwendener2010estimation}, multiple types of relations between financial TS have been uncovered. To test whether this prior knowledge has predictive capability, we propose that the relationships between the dynamic processes of multiple TS be given as additional prior inputs \(W \in \mathbb{R}_{+}^{n \times R \times n}\) to the model, as in \cite{ziat2017spatio}. We will first formalize how relationships between series are incorporated into the model and how this allows us, "virtually", to have a high number of training samples. Then, we will describe two extensions of this approach. The first extension allows the strength of these relations to be weighted, and the second allows the model to learn these relations directly without any prior information.

Relationships between the dynamic processes of \(n\) TS are incorporated via a tensor \(W \in \mathbb{R}_+^{n \times R \times n}\), where \(R\) is the number of relation types given as prior.In the following discussion, each sequence is indexed by \(X_{t,i}\), while the corresponding hidden state is represented by \(Z_{t,i}\). \(X_{t,i}\) correspond to the particular observation of the $i$\textsuperscript{iem} TS at time $t$ and \(Z_{t,i}\) is its corresponding hidden state. We formulate that, at time \(t\), \(Z_{t+1,i}\) depends on its own latent representation (intradependency) and on the representations of other series (interdependency).

Intradependency is modeled through a linear mapping \(\Theta^{(0)} \in \mathbb{R}^{n \times n}\). Interdependency is modeled with one transition matrix $\Theta^{(r)} \in \mathbb{R}^{n\times n}$ for each possible type of relation $r \in R$. $\Theta^{(r)}$ learns the relationship between each TS by applying a linear combination between neighboring TS and we denote $\Theta^{(r)}_i$ as the linear combination learned for relation $r$ of the $i$\textsuperscript{iem} series. We denote all the transition matrices by \(\Theta^{(R)} \in \mathbb{R}^{R \times n \times n}\) and $W_i^{(r)}$ as the relation given as prior between the $i$\textsuperscript{iem} series and other neighboring TS. To evaluate \(Z_{t+1}\), we compute the matrix product between the latent space \(Z_{t}\) and its dependencies (\(\Theta^{(0)}, \Theta^{(R)}\)) as in  Eq.~\ref{eq:dynamicfunctionformalized}. The decoder follows along, using \(Z_{t,i}\) as inputs, and computes the expected variation as in Eq.~\ref{eq:decoderfunctionformalized}. \(h_g, h_d\) are the respective activation functions of \(g(Z;\textbf{W}_g)\) and \(d(Z;\textbf{W}_g)\).

\begin{equation}
 \label{eq:dynamicfunctionformalized}
\begin{split}
     \tilde{Z}_{t+1, i} = {}& g(Z_{{t-k:t}, i}; \textbf{W}_g, \Theta)\\  ={} & h_g(actm_g(Z_t\Theta_i^{(0)}  + \sum_{r \in R} W_i^{(r)} Z_t\Theta_i^{(r)}); \textbf{W}_g) 
\end{split}
\end{equation}
\begin{equation}
 \label{eq:decoderfunctionformalized}
\begin{split}
    \tilde{X}_{t+1,i} = {X}_{t,i} + d(Z_{t,i}; \textbf{W}_d) = h_d(Z_{t, i}; \textbf{W}_d)
\end{split}
\end{equation}

Note that \(Z_t\) is shared among all series with respect to \(g(Z; \textbf{W}_g)\), but the representation of each series is disentangled explicitly by means of \(W\); i.e., \(d(Z;\textbf{W}_d)\) takes as input \(Z_{t,i}\), the hidden factor of the \(i\)\textsuperscript{th} TS. Doing this has two advantages:
\begin{enumerate*}[label={(\arabic*})]
\item \(g(Z;\textbf{W}_g)\) can forecast \(\tilde{Z}_{t+1}\) with fewer regressors.
\item It "virtually" increases the number of training samples, as we can use time and positional coordinates to make \(T \times n\) fixed-size training samples instead of handling TS as sequential data. With respect to each disentangled latent state, the correlation existing between the latent states of two TS would indicate that our model estimates that the TS follows similar trajectories despite $W$ specifying that they are or are not correlated which is possible in part due to $\Theta^{(R)}$.
\end{enumerate*}

\subsection{Model Extensions}
\label{Model extensions}

The two possible extensions proposed in \cite{ziat2017spatio} can also be applied to our model. We summarize the extensions here; readers are invited to refer to the original paper \cite{ziat2017spatio} for a more detailed explanation. The first extension, denoted by \texttt{STANN-R}, consists of adding a learned matrix of weights \(\Gamma^{r} \in \mathbb{R}_+^{n \times n}\) that can reduce the strength of relations given as prior. The second extension, denoted by \texttt{STANN-D}, consists of replacing \(W\) with \(\Gamma\) such that the model learns both the relational structure and the relation weights within \(\Gamma\). Applying the \texttt{STANN-R} or \texttt{STANN-D} extension formalizes Eq.~\ref{eq:dynamicfunctionformalized} as in Eq.~\ref{eq:dynamic_stnn_r} or Eq.~\ref{eq:dynamic_stnn_d}, respectively, where \(\odot\) signifies element-wise multiplication between two matrices:

\begin{equation}
\label{eq:dynamic_stnn_r}
\begin{split}
Z_{t+1, i} = g(Z_{t-k:t, i}; \textbf{W}_g, \Theta) = g( actm_g(Z_t\Theta_i^{(0)} + \sum_{r \in R} (\Gamma_i^{(r)} \odot W_i^{(r)}) Z_t\Theta_i^{(r)})
\end{split}
\end{equation}

\begin{equation}
\label{eq:dynamic_stnn_d}
\begin{split}
Z_{t+1, i} = g(Z_{t-k:t, i};\textbf{W}_g, \Theta) = g( actm_g(Z_t\Theta_i^{(0)} + \sum_{r \in R} \Gamma_i^{(r)}\Theta_i^{(r)})
\end{split}
\end{equation}

The optimization problem can thus be adjusted for \(\Gamma\), depending on whether the dynamic function is specified by Eq.~\ref{eq:dynamic_stnn_r} or Eq.~\ref{eq:dynamic_stnn_d}, and can be written as Eq.~\ref{eq:loss_stann}. \(|\Gamma|\) is a \(l_1\) regularizing term intended to sparsify \(\Gamma^{(r)}\); \(\gamma\) is a hyperparameter set to tune this term; and \(\lambda\) is a factor set to balance the relative importance of \(g(Z; \textbf{W}_g)\) and \(d(Z;\textbf{W}_d)\).

\begin{equation}
\label{eq:loss_stann}
\begin{split}
    d^* g^*, actm_g^*, \Theta^*, \Gamma^* = \operatorname*{argmin}_{d, Z, \Gamma,} {\dfrac{1}{T}}\sum_t \Delta(d(Z_{t}; \textbf{W}_d) + X_{t-1}, X_t) \\ + \gamma|\Gamma| + \lambda {\dfrac{1}{T}} \sum_{t=1}^{T-1} || Z_{t+1} - g(Z_{t-k:t}; \textbf{W}_g, \Theta)||^{2}
\end{split}
\end{equation}

\section{Experiments}
\label{Experiments}

\subsection{Datasets and Experimentation Procedure}
\label{Datasets}

\begin{figure*}[thbp]
\centering

\captionof{table}{Datasets for experimental evaluation} \label{tab:dataset}
\resizebox{.95\textwidth}{!}{%
\begin{tabular}{ccccccc}\toprule[1.5pt]
 Dataset &  \(\boldsymbol{T}\) &  \(\boldsymbol{n}\) &  Data type&  Time horizon &  \(\boldsymbol{\tau}\) & \# Runs per model \\\midrule
$\mathscr{D}_1$  & 2186  & 10  & daily adj. close & 1996/07/08 - 2007/08/22  & 21 & 100\\
$\mathscr{D}_2$  & 2000  & 69  & daily adj. close  & 2011/05/31
- 2019/05/10 & 21  & 54\\
\bottomrule[1.25pt]
\end {tabular}\par
\bigskip
}
\justify
\(\boldsymbol{T}\) is the total number of time points, \(\boldsymbol{n}\) the number of series, \(\tau\) the number of steps ahead to forecast and \# Runs the total number of evaluation runs made. For all datasets, we considered only the closing price \((m=1)\).
\end{figure*}

We report here the results of an experimental evaluation of our forecasting methods on two datasets: $\mathscr{D}_1=\texttt{Fasttrack}$ and $\mathscr{D}_2=\texttt{Fasttrack Extended}$. The two datasets, summarized in Table \ref{tab:dataset}, were obtained through FastTrack\footnote{\url{https://investorsfasttrack.com}}. They were selected for restraining the number of training samples and as representing respectively a low-data setting and a medium-data setting. Both datasets contain daily closing prices of U.S. MFs and ETFs traded on U.S. financial markets, each covering different types of asset classes including stocks, bonds, commodities, currencies and market indexes, or a proxy for a market index. Taken in combination, they cover 19 years of financial market activities and provide an overall view of the whole financial ecosystem. Each TS of these datasets represents the aggregation of multiple individual financial assets. In some of these TS, like \texttt{VFICX}, the aggregation of these individual TS is subject to vary over time with respect to management activities associated with these funds.

\begin{equation}
\label{eq:MASE}
    MASE(\tilde{X}, X) = \frac{1}{H}\sum_{i=1}^{H}\frac{|x_{T+i} - \tilde{x}_{T+i}|}{\frac{1}{T+H-s}\sum_{j=s+1}^{T+H}|x_j - x_{j-m}|}
\end{equation}

\begin{equation}
\label{eq:THEILU}
    THEILU(\tilde{X}, X) = \frac{1}{H}\sum_{i=1}^{H}\frac{\sqrt{|x_{T+i} - \tilde{x}_{T+i}|^2}}{\frac{1}{T+H-s}\sum_{j=s+1}^{T+H}\sqrt{|x_{T+i} - \tilde{x}_{T+i}|^2}}
\end{equation}

\begin{equation}
\label{eq:SDILAE}
    SDILATE(\tilde{X}, X, \tilde{X'}) = \frac{(\alpha) loss_{shape}(\tilde{X}, X) + (1-\alpha) loss_{time}(\tilde{X}, X)}{(\alpha) loss_{shape}(\tilde{X}', X) + (1-\alpha) loss_{time}(\tilde{X}', X)}
\end{equation}

\begin{equation}
\label{eq:MDA}
    MDA(\tilde{X}, X) = \frac{1}{N}\sum_{i=0}^\tau sign(\tilde{X}_{t:t+i}-X_{t-1})=sign(X_{t:t+i}-X_{t-1})
\end{equation}

To train each model, we carried out an evaluation on a rolling-forecasting-origin cross-validation, setting the number of time steps \(\tau\) to 21 days for simulating forecasting on a monthly basis. All models were trained on normalized TS using the interquartile range method. Produced forecasts were unscaled back to the original TS scales to measure the forecast's error. All DNN-based models were trained using stochastic gradient descent (SGD) with Adam \cite{kingma2014adam} and a learning rate scheduler \cite{LoshchilovH17}. The number of epochs, learning rate and other model hyperparameters, such as the optimal training window or the number of hidden layers and the number of hidden neurons for DNN-based models, were determined by a Bayesian hyperparameter search \cite{balandat2019botorch}.

\subsection{Time series Models}
\label{Time series models}

For fairness of comparison, we considered only models that can forecast multivariate TS directly, with the exception of two baseline models. The models used are as follows:

\begin{enumerate}

\item \noindent\textbf{Naive:} A simple heuristic that assumes the \(\tau\) future steps will be the same as the last previously observed.

\item \noindent\textbf{AR:} A classical univariate autoregressive process in which each TS is forecasted individually. The prediction is a linear function of past $l$ lags.

\item \noindent\textbf{ARIMA:} An autoregressive integrated moving average model that forecasts each TS individually. Implementation of ARIMA was done with \cite{PMDARIMA} to automatize the selection of the best parameterization over the training set.

\item \noindent\textbf{LSTM:} A long short-term memory model that forecasts \(\tau\) steps ahead in an iterative fashion \cite{hochreiter1997long}. LSTM with hidden layers and the number of hidden neurons were considered. 

\item \noindent\textbf{LSTM-A:} The same model as LSTM but with an added softmax attention layer to weight the importance of each past latent state for forecasting the next step-ahead. 

\item \noindent\textbf{WaveNet:} A convolutional neural network using dilated causal convolutions \cite{van2016wavenet}. 

\item \noindent\textbf{1-BEATS:} A member of the neural basis expansion analysis for time series forecasting ensemble model presented in \cite{oreshkin2019n}.

\item \noindent\textbf{STNN:} The closest model to ours. STNN can be considered as a particular case of our model, i.e., our model with \(k=1\). The two extensions of STNN (STNN-R and STNN-D) \cite{ziat2017spatio} were also considered. The Pearson correlations between TS were computed over the training set to define \(W\). We used the same training strategy as for STANN, i.e., modeling the variation only and training the model end-to-end to establish a fair comparison between model architectures.

\item \noindent\textbf{STANN:} The model proposed in this paper. The two extensions presented in Section~\ref{Model extensions} were also considered. The extensions expressed in Eq.~\ref{eq:dynamic_stnn_r} and Eq.~\ref{eq:dynamic_stnn_d} are denoted by STANN-R and STANN-D, respectively. Pearson correlation was used to define \(W\).
\end{enumerate}

\begin{figure*}[htbp]
    \centering
    \captionof{table}{Average forecasting performance of tested models on the \texttt{Fasttrack} dataset}
    \label{tab:results_Fasttrack}
    \resizebox{.85\textwidth}{!}{%
    \begin{tabular}{@{}l|llll@{}}\toprule[1.5pt]
& \multicolumn{4}{l}{FAST TRACK} \\ \midrule
\textit{\textbf{Model:}} & \textit{\textbf{MASE}} & \textit{\textbf{THEILU}} & \textit{\textbf{sDILATE}} & \textit{\textbf{MDA}}\\
\midrule
Naive & $1.0000 \pm 0.0000$ & $1.0000 \pm 0.0000$ & $1.0000 \pm 0.0000$   & $0.0180 \pm 0.0149^{****}$  \\
AR  & $1.0707 \pm 0.1517^{****}$ & $1.0757 \pm 0.1577^{****}$ & $1.1819 \pm 0.3560^{****}$   & $0.5085 \pm 0.1469^{***}$  \\
ARIMA & $1.0030 \pm 0.1205^t$ & $1.0133 \pm 0.1204^t$ & $1.0412 \pm 0.2457^*$   & $\textbf{0.5817} \pm  \textbf{0.1834}$   \\
LSTM & $1.3399 \pm 0.6020^{****}$ & $1.3405 \pm 0.6332^{****}$ & $2.1941 \pm 2.5503^{****}$   & $0.4861 \pm 0.1624^{****}$  \\
LSTM-A & $ 1.5708 \pm 0.6607^{****}$ & $1.5088 \pm 0.6261^{****}$ & $2.6648 \pm 2.4887^{****}$   & $0.4355 \pm 0.1642^{****}$ \\
WaveNet & $1.5936 \pm 0.7655^{****}$ & $1.6093 \pm 0.8133^{****}$ & $3.2449 \pm 3.7111^{****}$   & $0.4844 \pm 0.1606^{****}$ \\
1-BEATS & $ 2.9653 \pm 1.3327^{****}$ & $2.8485 \pm 1.3271^{****}$ & $9.8578 \pm 9.2178^{****}$   & $0.4307 \pm 0.1490^{****}$ \\
STNN & $\textbf{0.9852} \pm \textbf{0.0693}$ & $\textbf{0.9920} \pm \textbf{0.0756}$ & $\textbf{0.9897} \pm \textbf{0.1484}$ & $\textbf{\underline{0.5942}} \pm \textbf{\underline{0.1816}}$ \\
STNN-R & $\textbf{0.9860} \pm \textbf{0.0785}$ & $0.9900 \pm 0.0791^*$ & $0.9863 \pm 0.1431^*$ & $0.5450 \pm 0.1965^*$ \\
STNN-D & $1.0812 \pm 0.2957^{***}$& $1.0808 \pm 0.2765^{**}$ & $1.2439 \pm 0.8131^{**}$ & $0.5585 \pm 0.1533^t$ \\
\midrule
STANN & $\textbf{\underline{0.9792}} \pm \textbf{\underline{0.1045}}$ & $\textbf{\underline{0.9828}} \pm \textbf{\underline{0.1114}}$ & $\textbf{0.9783} \pm \textbf{0.2174}$ & $\textbf{0.5363} \pm \textbf{0.1914}$ \\
STANN-R & $\textbf{0.9806} \pm \textbf{0.0784}$ & $\textbf{0.9863} \pm \textbf{0.0804}$ & $\textbf{0.9793} \pm \textbf{0.1562}$ & $\textbf{0.5864} \pm \textbf{0.1873}$ \\
STANN-D & $\textbf{0.9864} \pm \textbf{0.0381}$ & $\textbf{0.9870} \pm \textbf{0.0374}$ & $\textbf{\underline{0.9756}} \pm \textbf{\underline{0.0707}}$ & $0.5642 \pm 0.1956^t$ \\
\bottomrule   
\end{tabular} 
}
\justify
Averaged forecasting results of the 21-day multivariate trajectory forecasts for both datasets. Boldface indicates the best methods who was determined by the Wilcoxon signed-rank test with significance level of $p-value < 0.10$. We also indicate the statistical significance of the difference from the best-performing model on the associated metric (t: p$\leq$ 0.10; *: p$\leq$ 0.05; **: p$\leq$0.01; ***: p$\leq$0.001; ****: p$\leq$0.0001). Underlining is used to indicate the best-performing model, comparing the significance level on all metrics.
\end{figure*}{}

\begin{figure*}[htp]
    \centering
    \captionof{table}{Average forecasting performance of tested models on the \texttt{Fasttrack extended} datasets}
    \label{tab:results_Fasttrack2}
    \resizebox{.85\textwidth}{!}{%
    \begin{tabular}{@{}l|llll@{}}\toprule[1.5pt]
& \multicolumn{4}{l}{FAST TRACK EXTENDED} \\ \midrule
\textit{\textbf{Model:}} & \textit{\textbf{MASE}} & \textit{\textbf{THEILU}} & \textit{\textbf{sDILATE}} & \textit{\textbf{MDA}}\\
\midrule
Naive &  $1.0000 \pm 0.0000$         & $1.0000 \pm 0.0000$         & $1.0000 \pm 0.0000$     & $0.0128 \pm 0.0082^{****}$ \\
AR  & $1.0337 \pm 0.0844^{****}$    & $1.0306 \pm 0.1033^{***}$      & $1.0723 \pm 0.2109^{***}$ & $0.4788 \pm 0.0955^*$ \\
ARIMA & $1.0011 \pm 0.0945^*$         & $1.0008 \pm 0.1193^t$        & $1.0156 \pm 0.2373^t$  & $0.2748 \pm 0.1222^{****}$ \\
LSTM & $1.2543 \pm 0.3020^{****}$    & $1.2311 \pm 0.3002^{****}$     & $1.6041 \pm 0.8031^{****}$  & $0.4821 \pm 0.1426^t$ \\
LSTM-A & $ 1.3940 \pm 0.3900^{****}$ & $1.3410 \pm 0.3713^{****}$ & $1.9345 \pm 1.1580^{****}$   & $0.4841 \pm 0.1357^{t}$ \\
WaveNet & $1.3988 \pm 0.5445^{****}$    & $1.4071 \pm 0.5930^{****}$     & $2.3042 \pm 2.4721^{****}$  & $0.4864 \pm 0.1472^*$ \\
1-BEATS & $ 2.9836 \pm 1.1605^{****}$ & $2.7788 \pm 1.1065^{****}$ & $8.7333 \pm 6.5840^{****}$   & $0.4565 \pm 0.1449^{****}$ \\
STNN & $\textbf{1.0020} \pm \textbf{0.1536}$ & $\textbf{0.9959} \pm \textbf{0.1591}$ & $\textbf{1.0165} \pm \textbf{0.3354}$ & $\textbf{0.5259} \pm \textbf{0.1822}$ \\
STNN-R & $\textbf{1.0122} \pm \textbf{0.1707}$ & $\textbf{1.0047} \pm \textbf{0.1698}$ &  $\textbf{1.0369} \pm \textbf{0.3687}$ & $\textbf{0.5241} \pm \textbf{0.1693}$ \\
STNN-D & $\textbf{0.9814} \pm \textbf{0.1147}$ & $\textbf{0.9791} \pm \textbf{0.1255}$ & $\textbf{0.9743} \pm \textbf{0.2495}$ & $\textbf{0.5401} \pm \textbf{0.2052}$ \\
\midrule
STANN & $\textbf{0.9832} \pm \textbf{0.1023}$ & $\textbf{0.9814} \pm \textbf{0.1084}$ & $\textbf{0.9750} \pm \textbf{0.2148}$ & $\textbf{0.5360} \pm \textbf{0.2030}$ \\
STANN-R & $\textbf{0.9836} \pm \textbf{0.1026}$ & $\textbf{0.9816} \pm \textbf{0.1098}$ & $\textbf{0.9755} \pm \textbf{0.2189}$ & $\textbf{0.5401} \pm \textbf{0.2051}$ \\
STANN-D & $\textbf{\underline{0.9795}} \pm \textbf{\underline{0.1016}}$ & $\textbf{\underline{0.9785}} \pm \textbf{\underline{0.1096}}$ & $\textbf{\underline{0.9694}} \pm \textbf{\underline{0.2176}}$ & $\textbf{\underline{0.5406}} \pm \textbf{\underline{0.2055}}$ \\
\bottomrule   
\end{tabular} 
}
\justify
Averaged forecasting results of the 21 days multivariate trajectory forecasts. We highlight the best methods in bold.
\end{figure*}{}

\subsection{Forecasting Performance}
Our experimental results are summarized in Table~\ref{tab:results_Fasttrack} and Table~\ref{tab:results_Fasttrack2}. First, we analyze the average performance of all the models and the statistical significance of the results obtained. Our model outperforms the DNN-based and statistical baselines in terms of all metrics on both datasets. The values of these metrics also indicate the superiority of the training framework of STNN and STANN as compared to the other models evaluated. By using the proposed attention mechanism and the TS decomposition approach of N-BEATS, STANN improves on the performance of its base model (STNN).

The augmentation trick used in the STNN and STANN models, presented in Eq.~\ref{eq:dynamicfunctionformalized} and Eq.~\ref{eq:decoderfunctionformalized}, is the largest contributing factor behind these results. By exploiting prior knowledge on the relation of these TS, STANN and STNN enhance their ability to forecast TS by "virtually" increasing the number of training samples despite using a shared latent state like LSTM and Wavenet. These results are very promising, considering that
\begin{enumerate*}[label={(\arabic*})]
    \item our approach achieved such results using a relatively small number of TS;
    \item it was trained solely using historical prices.
\end{enumerate*}
It is not surprising that DNN-based models (WAVENET, LSTM, LSTM-A, 1-BEATS) underperform compared to statistical baselines when trained in this setting, given their large parameterization and the small number of training samples at their disposal. Our results show that the augmentation trick of STNN and STANN appears to be a solution to the lack of training samples when such models are trained in a multivariate setting. We point out that, contrary to \cite{borovykh2018dilated}, we did not achieve similar MASE for the one-step-ahead forecast. We observe that during model training, we achieved similar results but the accuracy quickly dropped after the first 5 steps ahead in the first few epochs to yield a better overall forecast on the whole trajectory. Hence, there appears to be a trade-off between short-term forecast accuracy and the longer-term forecast accuracy when optimizing DNN-based models.

\begin{figure}[ht]
\centering
Comparison between STANN-D and STNN-D forecast errors for \texttt{Fasttrack extended} dataset\\
\includegraphics[width =0.82\textwidth]{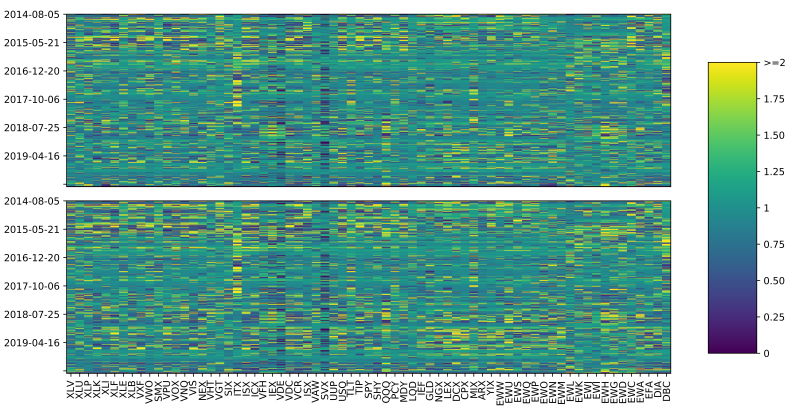}
\caption{Concatenation of the 21 daily return forecasts of STANN-D (top) and STNN-D (bottom). The absolute scaled error per series is presented.}
\label{fig:stann-d}
\end{figure}

\begin{figure}[bt]
    \centering
    Comparison between the evalauted models the \texttt{Fasttrack extended} dataset\\
    \includegraphics[width =0.82\textwidth]{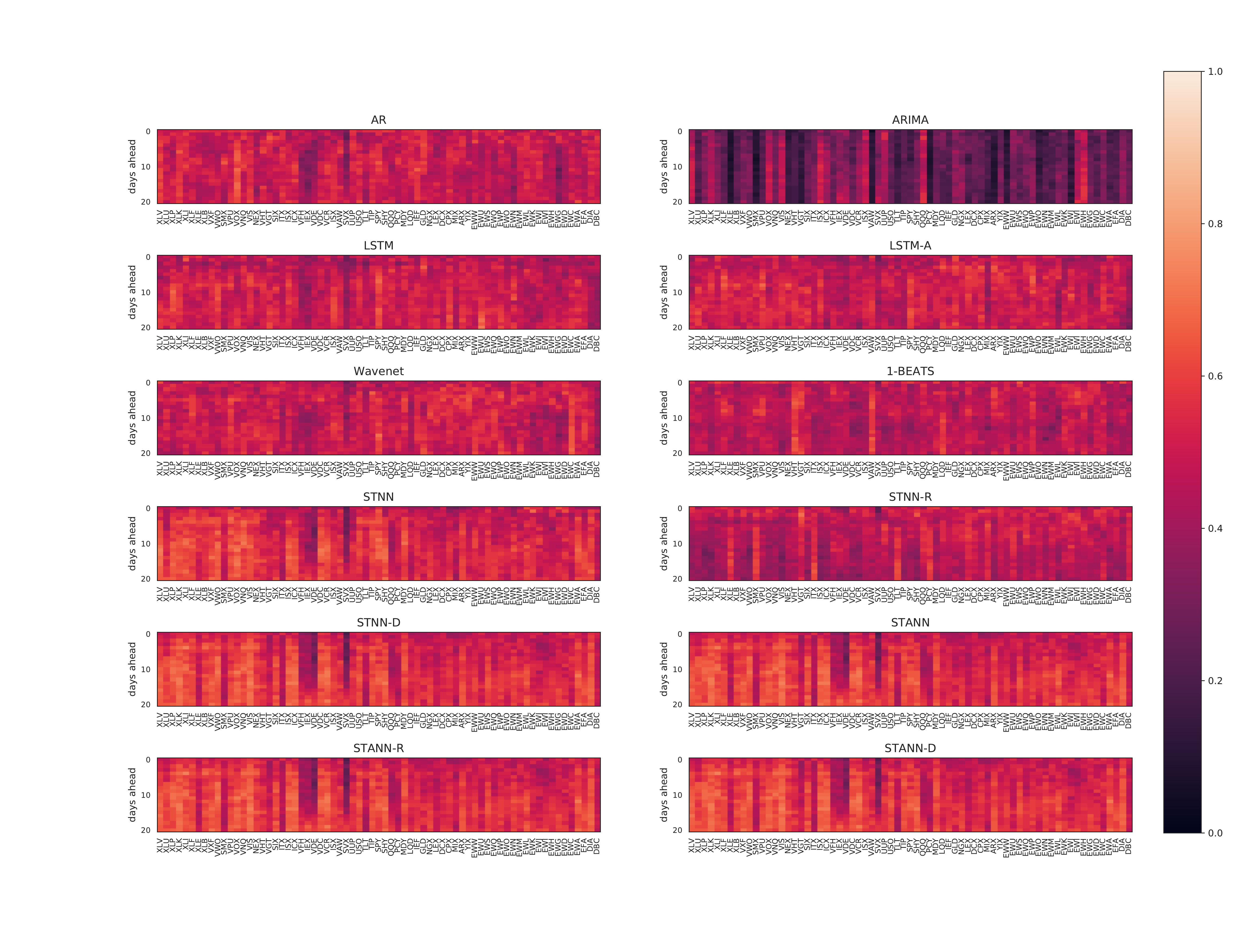}
    \caption{Directional accuracy for each $t$ steps ahead of the evaluated models. The darker the region, the less precise the model is at forecasting the trend of the trajectory of this particular asset.}
    \label{fig:mda_accuracy}
\end{figure}

We can qualitatively compare our models by plotting the absolute scaled error of the individual point forecast (IPF), i.e., $\frac{|x_{T+i} - \tilde{x}_{T+i}|}{\frac{1}{T+H-m}\sum_{j=m+1}^{T+H}|x_j - x_{j-m}|}$, for all the TS forecast, and comparing where our model fails. We observe that our approach is relatively consistent at forecasting the trajectory of each asset (Fig.~\ref{fig:stann-d}), but the majority of the residuals appear to occur in an epistemic fashion, i.e., the TS forecasting difficulty varies over time. Our proposed attention mechanism slightly increases the forecast accuracy in these episodes of forecast instability over its base model, which explains the majority of the additional average gain in accuracy. 

By looking at the ability of the evaluated models to predict the trend of the out-of-sample trajectory, we observe in Fig.~\ref{fig:mda_accuracy} that the STANN- and STNN-based models outperform baseline models but the STANN-based models show less variance in their results when compared to the different extensions of the STNN-based models. Given that these models tend to forecast the appropriate trend more accurately, this explains why our proposed framework outperforms other baseline models.

\begin{figure}[htbp]
    \centering
    Autoregressive Order of STANN\\
    \includegraphics[width=0.75\textwidth]{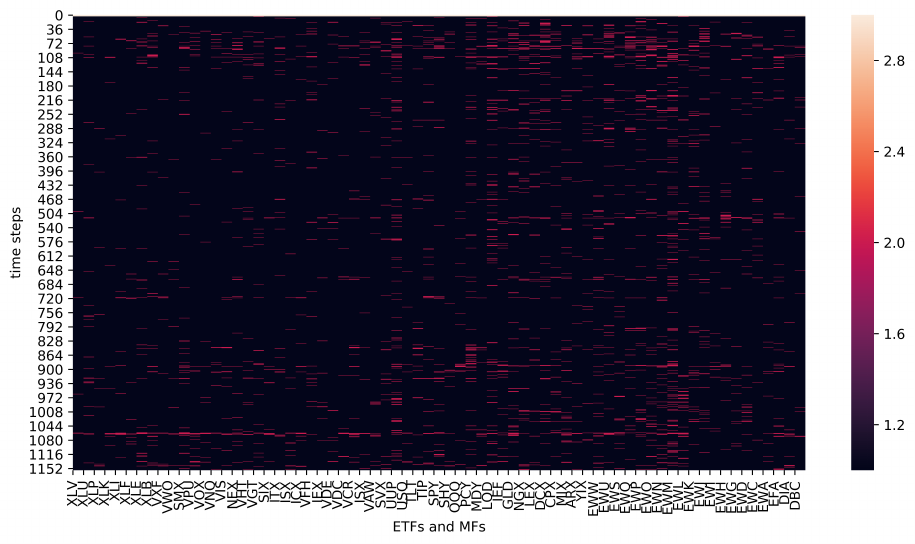}
    \caption{In-sample autoregressive order of a random instance of the STNN-R models taken from the \texttt{Fasttrack extended} evaluation. The left axis corresponds to past time steps used to train the model, with 0\textsuperscript{iem} steps being the furthest away from the prediction date; the bottom axis shows the securities forecast. Brighter colors indicate a higher autoregressive order for the time step in question.}
    \label{fig:ar_order}
\end{figure}

In Fig.~\ref{fig:ar_order}, we illustrate, showing the autoregressive order of a sample of the STANN-R model, that our model generates a dynamic process thanks to its attention mechanism. We can see that the model oscillates between AR(1) and AR(2) processes, although it remains more often at an AR(1) process. Interestingly, we can identify time spans for certain assets that are dominated by an AR(2) process and other regions dominated by an AR(1) process. This phenomenon is similar to what regime switching (RS) models \cite{hamilton1989new} enforce as prior when modeling the TS. Our approach differs in that we do not have to specify the number of regimes, nor the AR order \textit{a priori}. However, we notice that the choice of hyperparameters and model architecture can lead to vastly different results, with instances converging to either stationary or higher-AR-order solutions. Hence, further studies are needed to determine how this attention mechanism permits modeling of regime switches within its latent states. 

\begin{figure}[htbp]
\centering
Individual step-ahead MASE distribution of our approach on the \texttt{Fasttrack extended} dataset.\\
\includegraphics[width =0.85\textwidth]{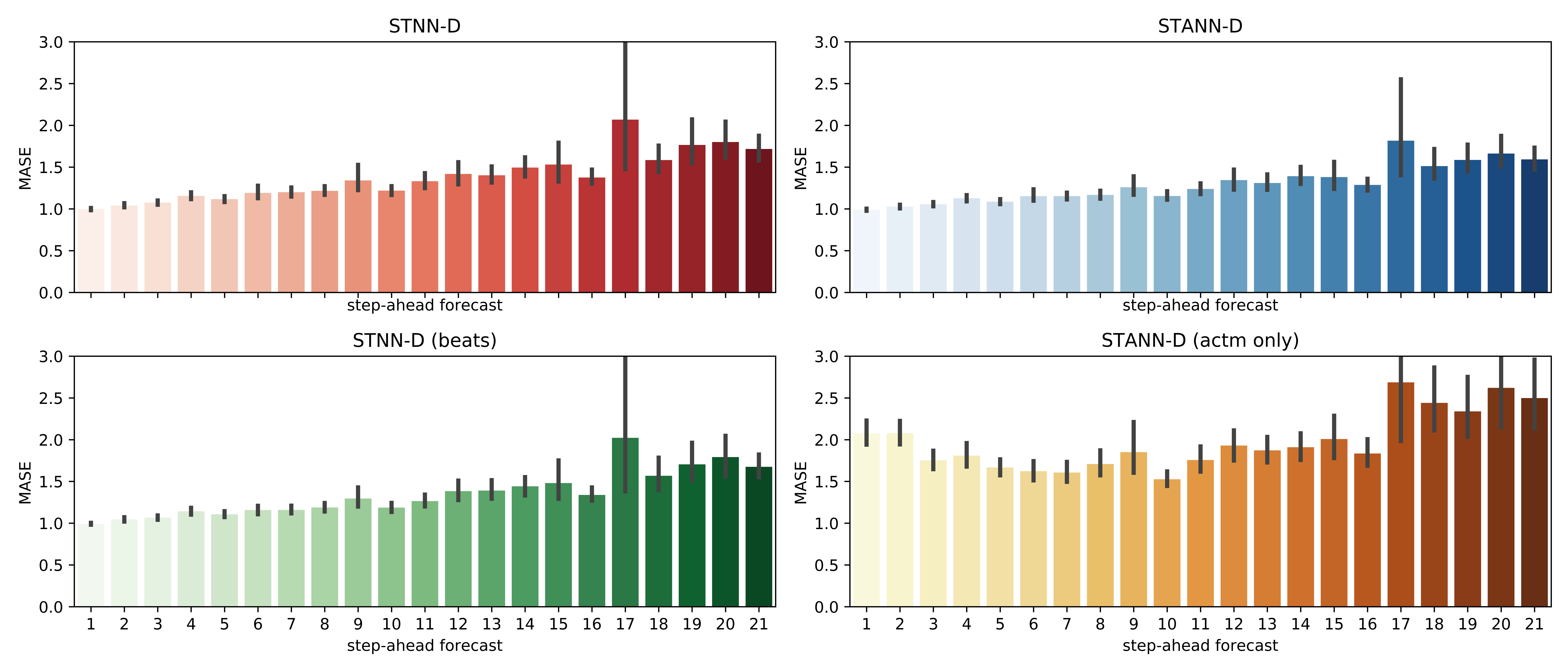}
\caption{An ablation study of STANN-D is presented. The median absolute scaled error for STANN-D is equal to 0.9671 and for STNN-D, to 0.9810}
\label{fig:kde}
\end{figure}

Finally, we also performed an ablation study to show the effect that the TS decomposition technique and the attention mechanism have on the model. To this end, we plotted the values of MASE metrics of each step-ahead forecast of model STNN-D and STANN-D with one of the two components removed. The plot, presented in Fig.~\ref{fig:kde}, shows the reference model (top left), our proposed model (top right), our model without ACTM (bottom left) and our model without the N-BEATS architecture (bottom right). The plot shows that using the attention mechanism alone without TS decomposition increases the overall forecast error but reduces the error propagation often found in recursive approaches. When combined with the TS decomposition architecture presented in \cite{oreshkin2019n}, we observe a significant error reduction for the last 14 days of the forecast trajectory. Regardless of the number of steps ahead forecast, our approach is significantly better than its base model at reducing IPF \(^{****}\).

\subsection{Added Value for Autonomous Decision Making}

We conducted some preliminary studies to demonstrate the utility of the proposed model to help autonomous decision making. Specifically, we show how the improved forecasting accuracy achieved by our model enhances the performance of autonomous trading strategies. The challenges of this question are threefold:
\begin{enumerate*}[label={(\arabic*})]
\item TS models are not decision systems in themselves.
\item We must rely on a trading strategy that considers forecast as a proxy to compare forecasting models. Given the myriad of strategies that exist, selecting one trading strategy over another is highly subjective and may lead to ambiguous results \cite{roll1978ambiguity}.
\item Each trading strategy has its own sensitivity and some trading strategies bypass the forecasting step by directly generating trade positions \cite{zhang2020deep}.
\end{enumerate*}
This last point is particularly important, as different trading strategies will produce different allocations depending on the same input data. Consequently, the sensitivity of trading strategies, and the state space on which they operate to establish trade signal plays a significant role in determining the excess return. This results in multiple issues that render it difficult to isolate the exact role that forecast accuracy has on various trading approaches, especially more sophisticated ones. It is necessary to point out that an extensive study of these issues is beyond the scope of this paper. However, we discuss briefly these issues and point out similarities and differences  that our approach has over existing ones.

One of the most popular frameworks for portfolio optimization, i.e., the traditional mean/variance framework \cite{rubinstein2002markowitz}, is a typical framework where the expected returns and variances are estimated and asset allocation decided based on an optimization procedure that aims to minimize a loss function. This approach is known to produce mixed results when exposed to noisy forecasts \cite{connor1997sensible}. Alternatively a simple strategy that generates buy and trade signals like in \cite{bao2017deep} can be made to translate the trajectory predicted in a trading action. The advantage of this approach is that they are model-agnostic, i.e., they do not depend on the model choice to apply the prediction and there exists correlation between having good estimates of the returns and the performance of a trading strategy. However, past profits can vanish very rapidly if the trading strategy makes a bad allocation at a bad time. Even if a TS model's accuracy is good on average, it suffices for the TS forecast to be \textit{bad} at the wrong moment to lose profits from past months. Hence, the timing of the forecasts plays a significant role in the performance of these strategies.

In comparison, other DNN-based strategies can be used to compare our TS model. These approaches generate directly trade signals within their architecture. This is either done in typical supervised setting like in \cite{passalis2019deep, tsantekidis2017forecasting} where a model aims to solve classification task with each class corresponding to a trade action. In this setting, a trade action must be labeled for each input data. Alternatively, RL approaches like \cite{zhang2019deeplob,deng2016deep,zhang2020deep} can be applied where a model generates trade actions based on their perceived state of their environment. These methods learn a policy which generates a probability distribution over possible trade actions based on a state space and are trained to maximize a reward function. This state space can be fabricated by a selection of different features as in \cite{zhang2020deep}. In this setting, we do not require labels for each input data, but we must specify a reward function for assessing the quality of the action chosen based on its environment. To compare the effect of TS models accuracy using either of these two approaches, TS forecast must be included within their input data.

However, additional issues arise from the use of these methods. As stated in \cite{zhang2020deep}, RL approaches can exhibit "slow learning and need a lot of samples to obtain an optimal policy." Similarly, this can also be the case of a DNN-based model depending on the model architecture. Unless we have at our disposal, the historical prediction for each time interval at all time points, using DNN and RL approaches is impractical as the amount of training samples to train these methods is diminished by the pace at which we can produce forecasts. Hence, for all DNN-based methods (LSTM, LSTM-A, WAVENET, STNN and STANN), one must resort to zero-shot forecasting, i.e., without explicit retraining on that target data, to produce these forecasts in a reasonable amount of time to justify the use of RL approaches. This, however, augments exposure to concepts drift effect that occurs when not retraining model \cite{vzliobaite2010learning}, which, in turns, tends to lower the average accuracy of the TS model \cite{oreshkin2020meta}. Hence, two additional factors can influence the performance of this approach: (1) sensibility to hyperparameters and model architecture of the trading strategy, (2) the number of training samples at our disposal where a forecast has been made. To prevent the leakage of these additional issues, we restrict our experimental analysis to two traditional approaches, which we describe below. 

\begin{table}[htbp]
Performance of various portfolio strategies built on the evaluated models \\
\resizebox{\textwidth}{!}{
\centering
\begin{tabular}{@{}l|llll|llll@{}}\toprule[1.5pt]
            & \multicolumn{4}{l}{FAST TRACK} & \multicolumn{4}{l}{FAST TRACK EXTENDED} \\ \midrule
            \textit{\textbf{Model:}} & \textit{\textbf{Sharpe}} & \textit{\textbf{Max             Drawdown}} & \textit{\textbf{mean return}} & \textit{\textbf{profit}} & \textit{\textbf{Sharpe}} & \textit{\textbf{Max             Drawdown}} & \textit{\textbf{mean return}} & \textit{\textbf{profit}}\\
            \midrule
            \multicolumn{9}{c}{\textbf{BL strategy}} \\
            \midrule
            AR & $0.37$ & $-2.41\%$ & $0.34\%$ & $39.87\%$ & $0.80$ & $-2.30\%$ & $0.24\%$ & $14.82\%$ \\
            ARIMA & $0.47$ & $-2.40\%$ & $0.35\%$ & $41.74\%$  & $0.97$ & $-2.47\%$ & $0.30\%$ & $18.79\%$ \\ 
            LSTM & $0.67$ & $-1.87\%$ & $0.39\%$ & $46.71\%$ & $0.82$ & $-2.84\%$ & $0.28\%$ & $17.27\%$ \\
            LSTM-A  & $0.49$ & $-2.23\%$ & $0.36\%$ & $42.62\%$ & $1.13$ & $-3.02\%$ & $0.35\%$ & $21.91\%$ \\
            WaveNet & $-0.02$ & $-48.52\%$ & $0.32\%$ & $30.58\%$ & $0.52$ & $-6.90\%$ & $0.27\%$ & $16.43\%$ \\
            1-BEATS & $-0.04$ & $-39.76\%$ & $0.30\%$ & $29.46\%$ & $0.05$ & $-10.02\%$ & $0.10\%$ & $5.19\%$ \\
            STNN & $0.49$ & $-2.21\%$ & $0.35\%$ & $41.95\%$ & $0.76$ & $-2.61\%$ & $0.22\%$ & $13.51\%$ \\
            STNN-R & $0.50$ & $-2.21\%$ & $0.35\%$ & $42.16\%$ & $1.11$ & $-4.13\%$ & $0.54\%$ & $35.70\%$ \\
            STNN-D & $0.48$ & $-2.13\%$ & $0.35\%$ & $41.72\%$ & $0.74$ & $-2.64\%$ & $0.22\%$ & $13.52\%$ \\
            \midrule
            STANN & $0.48$ & $-2.19\%$ & $0.35\%$ & $41.70\%$ & $0.73$ & $-2.57\%$ & $0.22\%$ & $13.26\%$  \\
            STANN-R & $0.47$ & $-2.20\%$ & $0.35\%$ & $41.67\%$ & $0.73$ & $-2.67\%$ & $0.22\%$ & $13.37\%$ \\
            STANN-D & $0.49$ & $-2.23\%$ & $0.35\%$ & $41.92\%$ & $0.72$ & $-2.66\%$ & $0.22\%$ & $13.22\%$ \\
            \midrule
            \textbf{Optimal} & $\textbf{3.97}$ & $\textbf{-1.92\%}$ & $\textbf{2.79\%}$ & $\textbf{1436.73\%}$ &
            $\textbf{2.89}$ & $\textbf{-5.30\%}$ & $\textbf{2.27\%}$ & $\textbf{263.39\%}$ \\
            \midrule
            \multicolumn{9}{c}{\textbf{Simple strategy}} \\
            \midrule
            AR & $-0.26$ & $-20.70\%$ & $0.18\%$ & $16.60\%$ & $0.38$ & $-16.56\%$ & $0.37\%$ & $21.59\%$ \\
            ARIMA & $0.47$ & $-11.58\%$ & $0.50\%$ & $62.22\%$ & $0.51$ & $-11.85\%$ & $0.40\%$ & $24.63\%$ \\ 
            LSTM & $0.15$ & $-29.69\%$ & $0.44\%$ & $47.40\%$ & $\textbf{0.76}$ & $-21.08\%$ & $\textbf{0.58\%}$ & $\textbf{37.39\%}$ \\
            LSTM-A & $-0.06$ & $-35.53\%$ & $0.28\%$ & $27.54\%$& $\textbf{0.53}$ & $-17.18\%$ & $\textbf{0.52\%}$ & $\textbf{31.51\%}$ \\
            WaveNet & $-0.02$ & $-48.52\%$ & $0.32\%$ & $30.58\%$ & $0.41$ & $-15.28\%$ & $0.35\%$ & $20.35\%$ \\
            1-BEATS & $-0.04$ & $-39.76\%$ & $0.30\%$ & $29.46\%$ & $0.21$ & $-17.41\%$ & $0.31\%$ & $16.18\%$ \\
            \textbf{STNN} & $0.49$ & $-11.47\%$ & $0.51\%$ & $63.55\%$  & $0.40$ & $-15.64\%$ & $0.37\%$ & $21.52\%$  \\
            STNN-R & $\textbf{0.54}$ & $-12.26\%$ & $0.50\%$ & $62.24\%$  & $\textbf{0.70}$ & $\textbf{-0.38\%}$ & $0.18\%$ & $10.95\%$ \\
            \textbf{STNN-D} & $\textbf{0.75}$ & $\textbf{-6.43\%}$ & $\textbf{0.61\%}$ & $\textbf{81.36\%}$ & $0.48$ & $\textbf{-10.73\%}$ & $0.41\%$ & $24.70\%$ \\
            \midrule
            \textbf{STANN} & $\textbf{0.52}$ & $-19.59\%$ & $\textbf{0.56\%}$ & $\textbf{70.75\%}$ & $0.50$ & $\textbf{-10.86\%}$ & $\textbf{0.43\%}$ & $\textbf{25.82\%}$  \\
            STANN-R & $0.45$ & $-12.07\%$ & $0.49\%$ & $60.64\%$ & $0.48$ & $\textbf{-10.73\%}$ & $0.41\%$ & $24.70\%$ \\
            \textbf{STANN-D} & $\textbf{0.66}$ & $\textbf{-8.99\%}$ & $\textbf{0.55}\%$ & $\textbf{71.13\%}$ & $0.48$ & $\textbf{-10.73\%}$ & $0.41\%$ & $24.64\%$ \\
            \midrule
            Equal Weight & $0.47$ & $-11.58\%$ & $0.50\%$ & $62.22\%$ & $0.48$ & $-10.73\%$ & $0.41\%$ & $24.70\%$ \\
            \bottomrule
\end{tabular}
}
\caption{Portfolio performance metrics of the three types of strategy, evaluated on the same time horizon utilized in Table~\ref{tab:dataset}. The best strategy is highlighted in bold.}
\label{tab:results_Fasttrack_extended_portfolio}
\end{table}

\begin{figure}[htbp]
\centering
Portfolio excess returns on the \texttt{FASTRACK} dataset
\begin{minipage}{.5\textwidth}
  \centering
  \includegraphics[width=\linewidth]{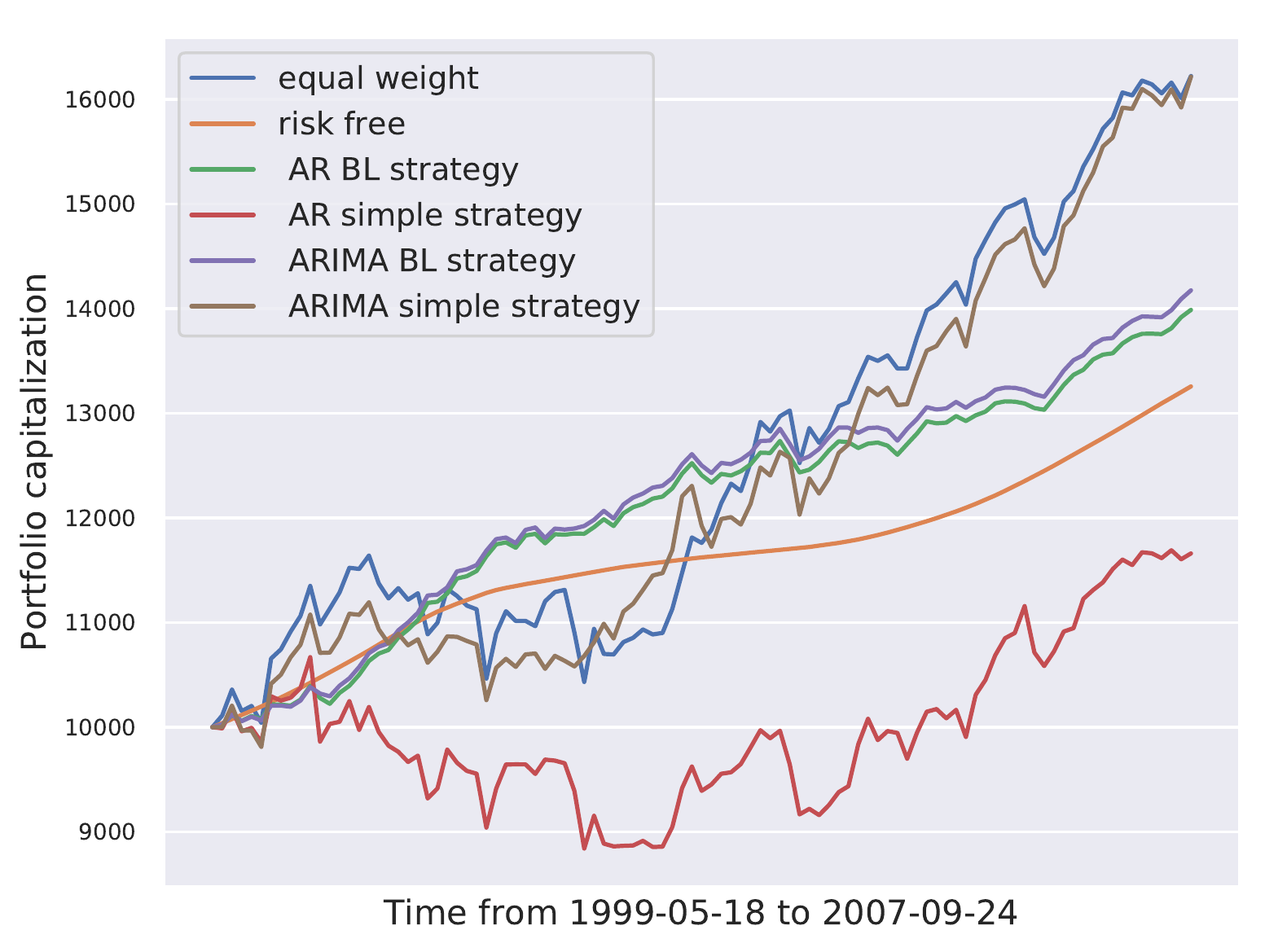}
\end{minipage}%
\begin{minipage}{.5\textwidth}
  \centering
  \includegraphics[width=\linewidth]{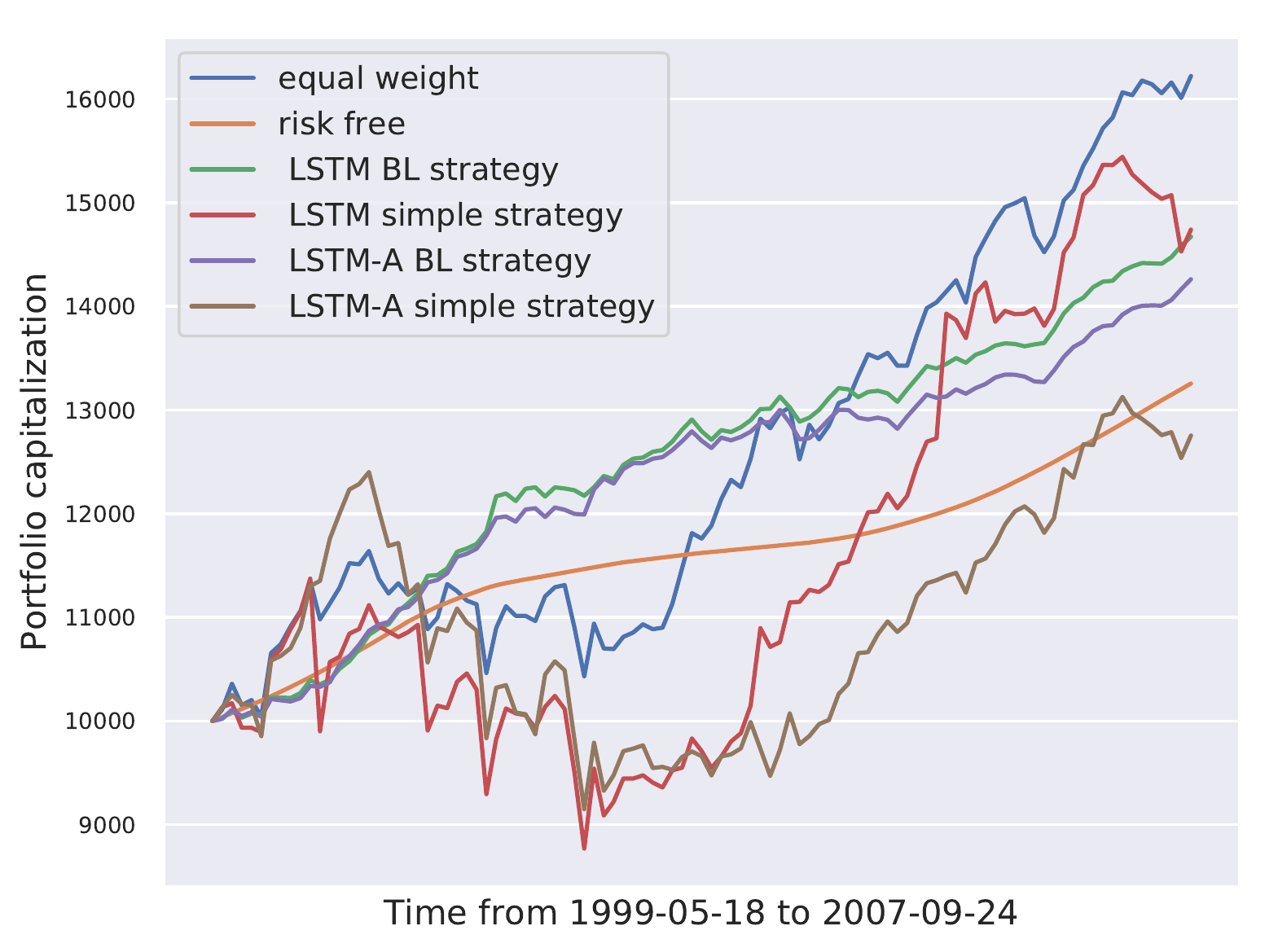}
\end{minipage}
\captionof{figure}{Portfolio performances based upon alternative statistical baselines (left) and RNN-based models (right)}.
\label{fig:portfolio_1}
\end{figure}

\begin{figure}[htbp]
\centering
Portfolio excess returns on the \texttt{FASTRACK} dataset
\begin{minipage}{.5\textwidth}
  \centering
  \includegraphics[width=\linewidth]{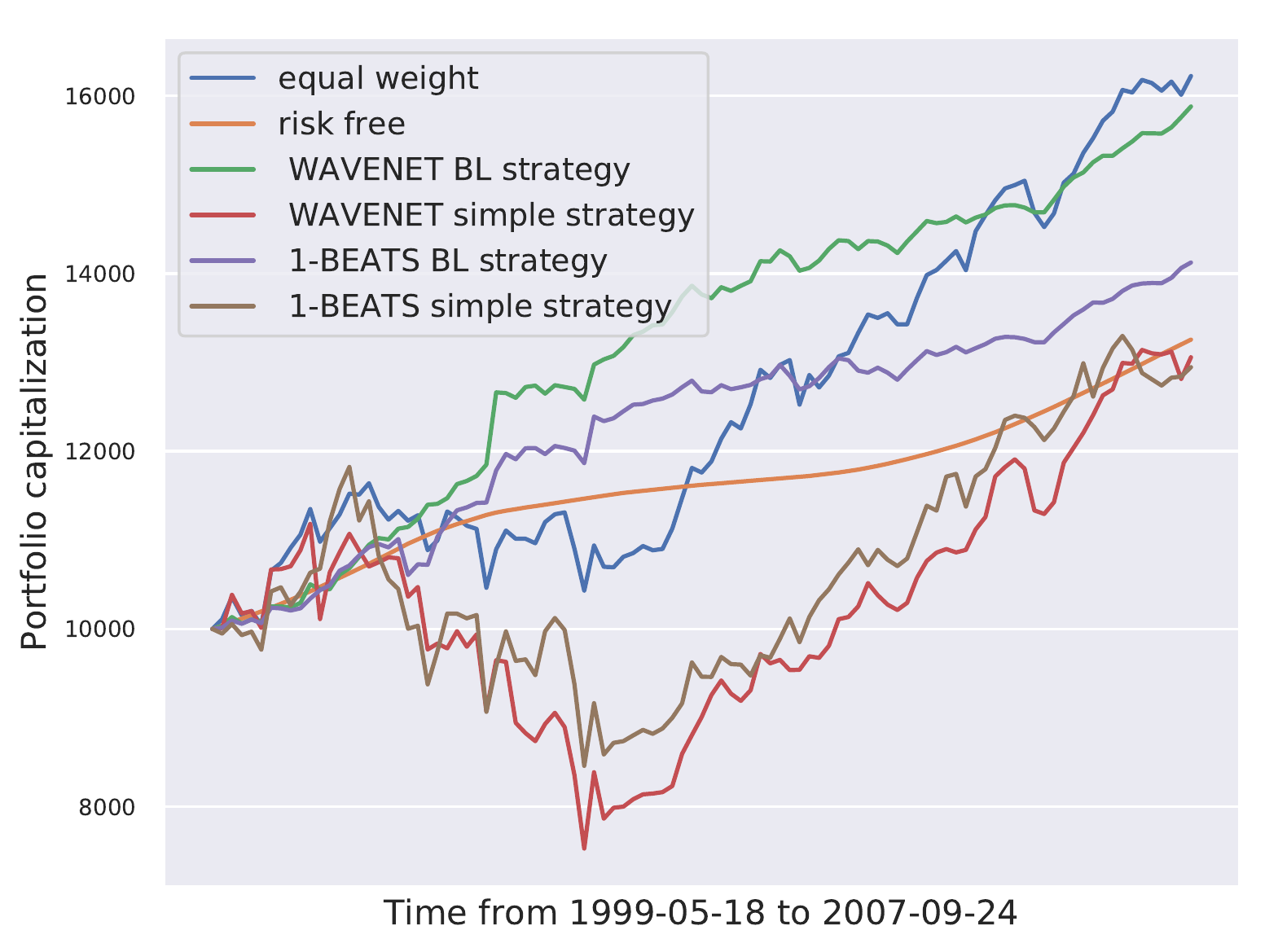}
\end{minipage}%
\begin{minipage}{.5\textwidth}
  \centering
  \includegraphics[width=\linewidth]{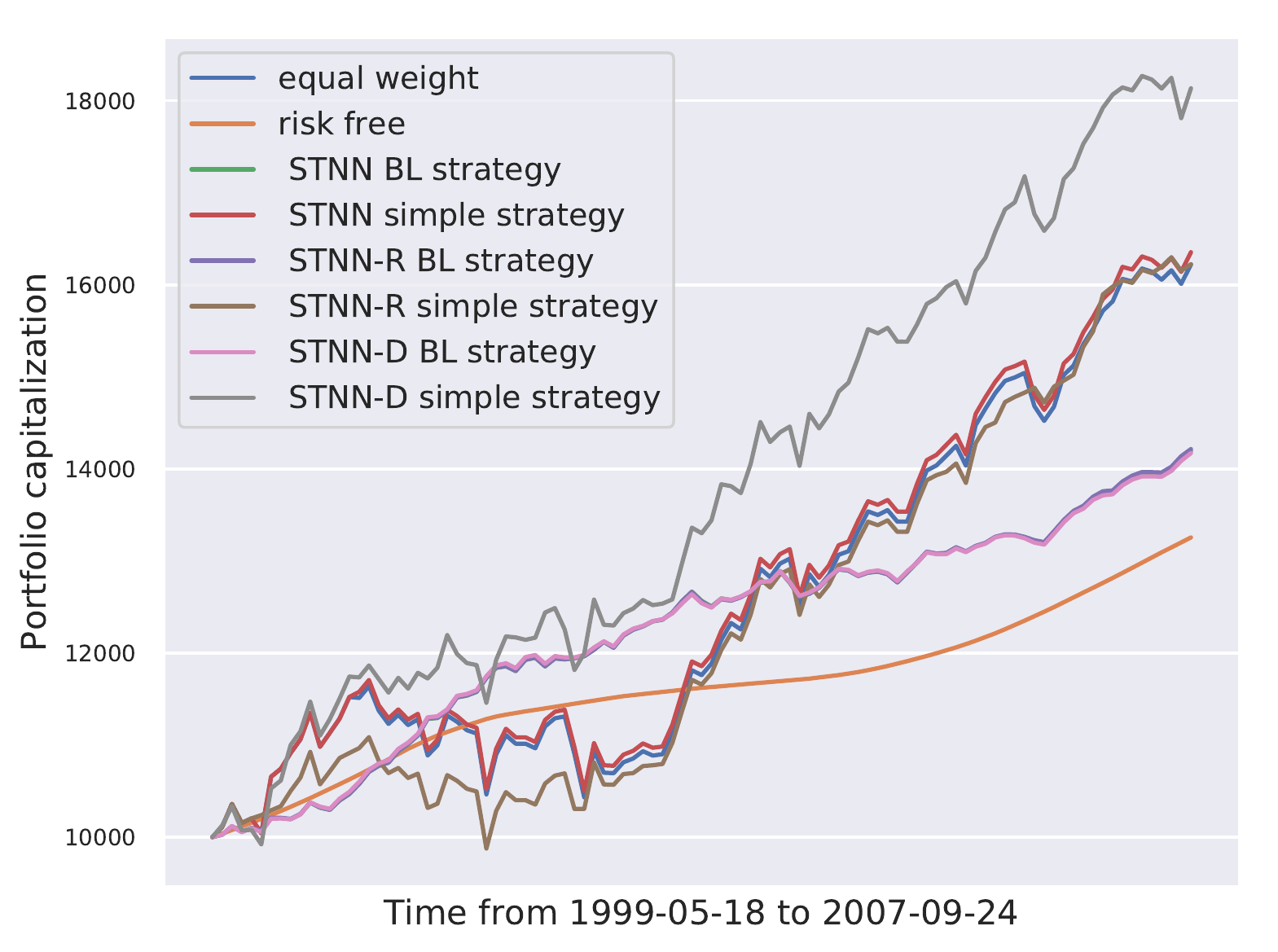}
\end{minipage}
\captionof{figure}{Portfolio performances based on alternative DNN models (left) and STNN-based models (right)}
\label{fig:portfolio_2}
\end{figure}

\begin{figure}[htbp]
    \centering
    Portfolio excess returns of \texttt{FASTRACK} dataset
    \includegraphics[width=0.70\linewidth]{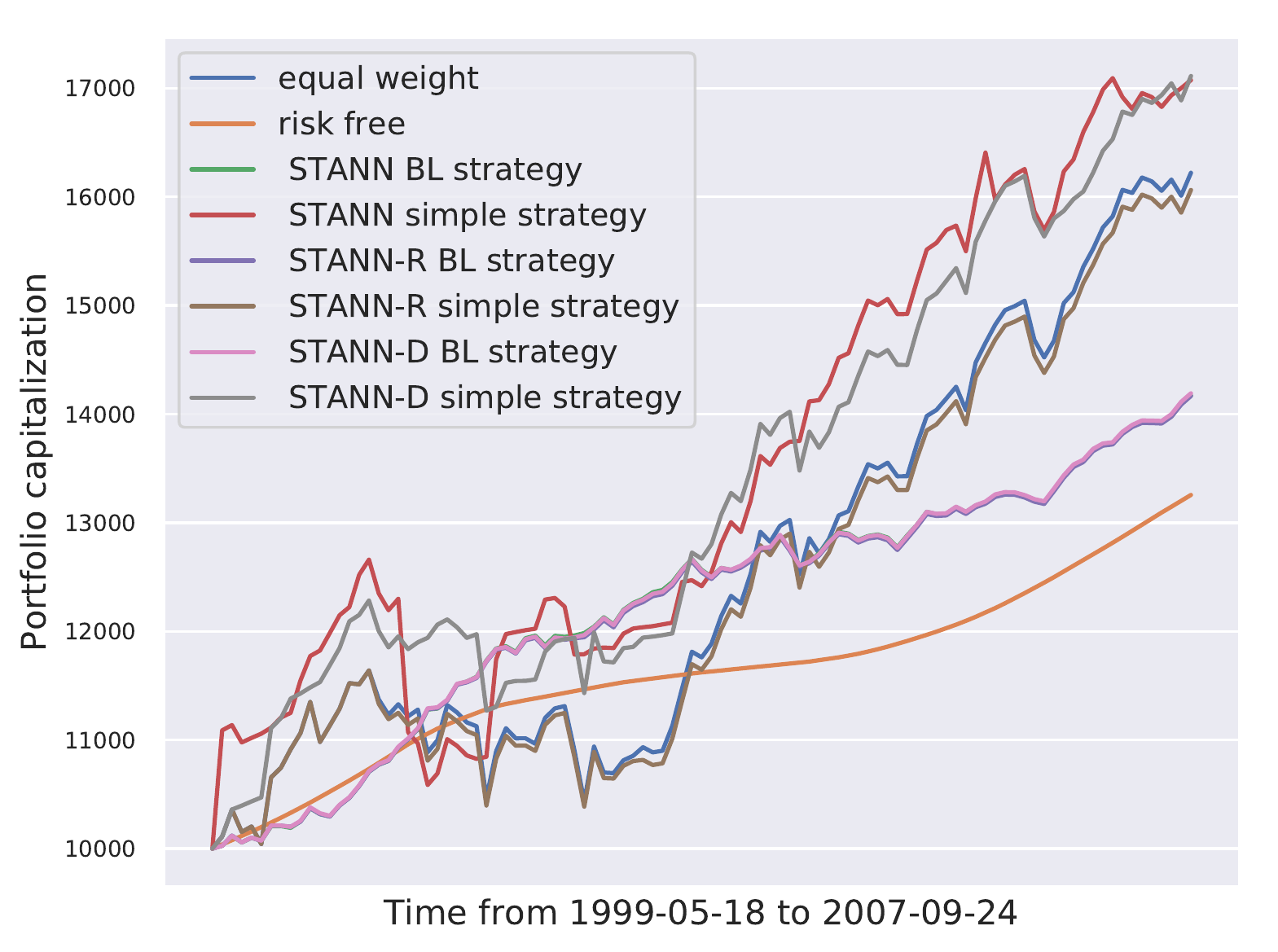}
    \caption{Portfolio performance based upon STANN-based models}
  \label{fig:portfolio_stann}
\end{figure}

In our experiments, we studied and observed relations between the accuracy of the TS forecast and the excess return of a strategy. Results for various financial metrics of two common strategies are presented in Table~\ref{tab:results_Fasttrack_extended_portfolio} on both datasets. Here, we include the annualized Sharpe ratio $\textit{sharpe}=\frac{E[r_{portfolio} - r_{risk free}])}{\sigma_{portfolio}} \times \sqrt{\mathscr{S}}$, where $r_{portfolio}$ is the return of the portfolio, $r_{riskfree}$ is the return of a risk-free strategy, $\sigma_{portfolio}$ is the standard deviation of the excess return of the portfolio and $\mathscr{S}=\tau/252$ is an annualization factor, with 252 being the average number of active market days per year. We also include the maximum drawdown, which is the maximum loss observed from a peak to a trough of a portfolio; the mean return per trade and the total profit after the observation periods. We also present the excess return graph of different portfolio strategies built on the evaluated models for the \texttt{FASTTRACK} dataset in Figs.~\ref{fig:portfolio_1}, ~\ref{fig:portfolio_2} and ~\ref{fig:portfolio_stann}.

We used two proxy trading strategies to carry out our evaluation.
\begin{enumerate}[label={(\arabic*})]
    \item \textit{Black-Litterman (BL) strategy}: The first is a simple, efficient frontier optimization trading strategy \cite{rubinstein2002markowitz} based on the BL allocation model \cite{black1990asset}, in which we maximized the Sharpe ratio. The expected returns are determined by the TS forecasts and implied market returns and the estimated risk was computed respectively by the in-sample returns and covariance.\footnote{A long-only portfolio constraint was added to ensure that the allocation weights remain between 0 and 1. In some instances this constraint is too restrictive and the convex optimization used in this approach fails. In such cases, we permit the optimization to consider the short position; i.e., allocation weights remain between -1 and 1, and do not consider the short position in our allocation.}
    \item \textit{Simple:} a simple trading strategy where one invests equally in each of the securities that the TS model predicted would increase in price. If the trend signal is negative and assets were allocated, we consider the forecast as a \textit{"sell"} signal. A positive forecast trend is interpreted as a \textit{"buy"} signal. The allocation weights were normalized according to the strength of the trend signal using a softmax function.
\end{enumerate}
We compare these strategies with the same "optimal" BL strategy where the expected returns and covariance matrix are known in advance to simulate how a perfect TS model would have performed for this strategy. We also include an equal-weight portfolio where we invest in each asset equally.

Allocations of assets were reconsidered on a 21-day basis to simulate how an autonomous trading strategy would change its portfolio allocation over time depending on the financial context. We used a \$10,000 initial portfolio value and bought whole security shares, with the leftover money being at a risk-free rate determined by the 30-day U.S. Treasury bill rate. For the \texttt{FASTTRACK} dataset, we used the U.S. 3-Month Treasury rate rather than the 1-month rate for the risk-free strategy, given that no data for the U.S. 1-Month Treasury rate before 2001-07-31 is available from the public and proprietary data sources at our disposal. Transaction fees were assumed to be null and dividends were accounted for within the adjusted closing price.

It can be seen that perfect predictions yield extremely significant returns for BL strategies on both datasets. However, although LSTM and LSTM-A are poor forecasters with respect to all metrics, the strategy built on them is among the top performers. Identifying the principal cause of this phenomenon is not a trivial matter. However, efficient frontier optimization methods are known to produce mixed results when the forecasts are too noisy \cite{connor1997sensible}. At this level of accuracy, none of the TS models evaluated are sufficiently robust by themselves to be considered in a BL strategy. However, we can observe that there are significant gains to be made if one could identify the condition where a TS forecaster helps a trading strategy, as each strategy overperforms the baselines at certain moments.

The performance of the simple strategy shows the importance of having a more accurate TS model for autonomous trading strategies. TS models that forecast more accurately, like the STNN and STANN-based approaches, are among the top performers on both datasets with respect to most metrics, especially when few TS are considered. When more TS are considered (\texttt{Fasttrack extended}), the simple strategy is too naive to select a meaningful subset of securities and will yield performance similar to an equal-weight portfolio. However, since STNN and STANN-based models forecast the trend more accurately, the maximum drawdowns of the simple strategy based on these models were much smaller compared to the LSTM and LSTM-A strategies, which made allocations over a smaller set of securities at a time. Hence, a simple naive strategy can perform relatively well within a curated set of TS using a better forecaster, but will not scale effectively when the number of assets considered is too large.

\section{Discussion}
\label{Conclusion}
This paper proposed a new self-supervised deep generative model (STANN) for forecasting multivariate TS conjointly, which explicitly models the interactions between TS. We introduced a novel attention-based mechanism that enhances the capability of any RNN based on the DFG framework. We showed how this attention-based mechanism increases the set of probability distributions that can be modeled by permitting modeling of non-stationary distributions. Incorporating this into our model, we presented one general approach and two extensions for considering interrelations between TS. We showed that when these interrelations are incorporated, we can fit these DNN-based models even where little training data exists. Experiments were performed on two financial datasets covering more than 19 years of market history. Our experiments indicate that STANN provides a more effective learning framework than either DNN-based approaches or statistical baselines. We showed that this class of models perform wells in both low- and medium-data settings and that our proposed attention mechanism helps improve forecasting performances over its base model. Finally we illustrated how the use of a forecaster improves autonomous trading strategies.

We would like to emphasize the limited understanding of the relation between our model  effectiveness and the selection of HPs. Indeed, a mis-selection of HPs can have a great impact on a model's performance, potentially hindering its application at a large scale. Hence, we advocate the pursuit of future work to enlarge our theoretical understanding of this class of models, as well as testing to determine whether similar results can be achieved at larger scales and for other TS settings.

\section*{Acknowledgements}
This work was supported in part by NSERC CRD - Quebec Prompt - Laplace Insights - EAM - joint program under the grants CRDPJ 537461-18 and 114-IA-Wang-DRC 2019 to Dr. S. Wang, by Mitacs Acceleration under Grant IT13429 to Dr. S. Wang and P. Chatigny and by FQRNT under Grant B2 270188 to P. Chatigny.






\bibliographystyle{elsarticle-num-names}
\bibliography{sample.bib}







\end{document}